\documentclass[letterpaper, 10 pt, conference]{ieeeconf}  

\IEEEoverridecommandlockouts                              
\usepackage{graphics} 
\usepackage{epsfig} 
\usepackage[hang,scriptsize]{subfigure}

\title{\LARGE \bf
Multi-Class Detection and Segmentation of Objects in Depth
}

\author{Cheng Zhang \and Hedvig Kjellstr{\"o}m
\thanks{This work was supported by the Swedish Research Council.}
\thanks{The authors are with the Computer Vision and Active Perception Lab and the Centre for Autonomous Systems, CSC, KTH, Sweden. {\tt\small chengz,hedvig@kth.se}}%
}

\begin{document}

\maketitle
\thispagestyle{empty}
\pagestyle{empty}

\begin{abstract}



The quality of life of many people could be improved by autonomous humanoid robots in the home. To function in the human world, a humanoid household robot must be able to locate itself and perceive the environment like a human; scene perception, object detection and segmentation, and object spatial localization in 3D are fundamental capabilities for such humanoid robots.
This paper presents a 3D multi-class object detection and segmentation method. The contributions are twofold. Firstly, we present a multi-class detection method, where a minimal joint codebook is learned in a principled manner. Secondly, we incorporate depth information using RGB-D imagery, which increases the robustness of the method and gives the 3D location of objects -- necessary since the robot reasons in 3D space.
Experiments show that the multi-class extension improves the detection efficiency with respect to the number of classes and the depth extension improves the detection robustness and give sufficient natural 3D location of the objects.
\end{abstract}

\section{Introduction}
\label{sec:intro}

Imagine a scenario where a humanoid household robot is setting up a table for dinner: First, it needs to locate itself in kitchen. 
It then has to detect objects, like forks and knifes, localize and grasp them. After that it needs to detect the location of the table and what is on the table, to perform the task of putting the tableware in the right location.
We can see that for a humanoid household robot to perform this kind of daily tasks, it needs to localize itself in the world, recognize and detect objects, localize them in 3D, and manipulate them. 

Simultaneous localization and mapping (SLAM) is supported by scene classification (room classification) which can be done in terms of detecting objects in the room \cite{torralba03,pronobis10}.
Scene classification can also be employed to guide object detection and object search \cite{aydemir11,forssen08}.
In order to perform manipulation, 
detecting objects and segmenting objects in 3D serves as the preprocessing step \cite{Marton:2011}.

\begin{figure}[t]
\centerline{\includegraphics[width=\linewidth]{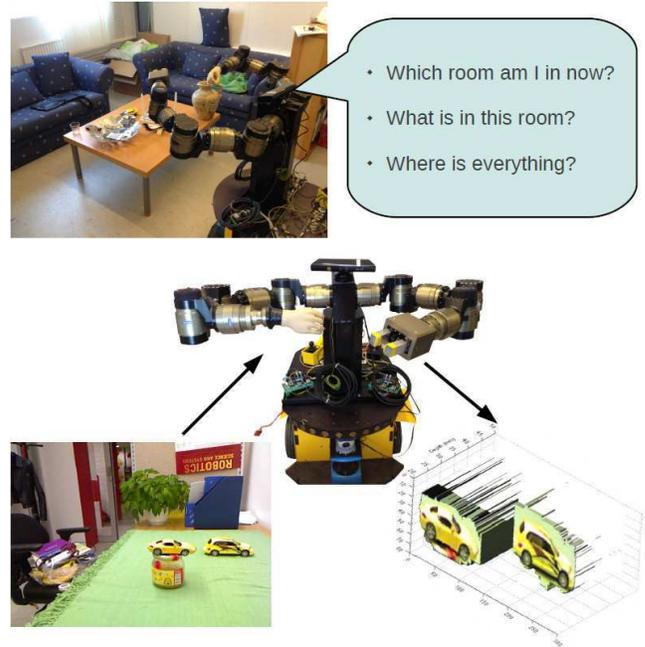}}
\vspace{-2mm}
\caption{An autonomous humanoid robot, which is used to do daily tasks to assist human, must be able to locate itself through scene perception,  detect the 
target object classes which would be manipulated with, and know the objects location to guide the manipulation.  
This paper presents a method for interleaved detection, segmentation, and 3D location of previously unseen objects from a range of 
known object classes using RGB-D imagery, which can be used for robot perception.
\vspace{-2mm}}
\label{fig:intro}
\end{figure}

In this paper, we present
 a method for {\em simultaneous, interleaved detection, segmentation and 3D 
localization of previously unseen object instances} of known categories {\em in unknown environments} (see Figure \ref{fig:intro}). The detection and segmentation processes guide each other in a contextual manner \cite{Erdos01} and we exploit depth localization to constrain both detection and segmentation.

Detecting and segmenting previously unseen object instances of known classes is a long-term challenging problem. However, it is essential for many robotic applications \cite{bohg11,marton11}.
Due to its complexity, the problem is often constrained in different ways; either by detecting
object {\em instances}, seen before \cite{aydemir11}, 
 or by introducing attention mechanisms \cite{bjorkman-kragic10,forssen08}. For 3D unseen object segmentation, different amounts of knowledge about the environment is required, for example, background subtraction  \cite{1400815}, planar support \cite{bjorkman-kragic10}, environment maps \cite{Marton:2011} etc. We avoid such assumptions and perform interleaved detection and segmentation of previously unseen object instances in 3D.

We employ the approach of Leibe et al.~\cite{Erdos01} for interleaved object detection and segmentation,
 where visual words in the image are voting for the center and boundaries of regions with a specific object category in them.
 This method is described in Section \ref{sec:leibe}. The contributions of the present paper are 
two enhancements of the approach \cite{Erdos01}, which make the object detections applicable for robotic reasoning. 

Firstly, we introduce a {\em multi-class detector}, which uses a joint codebook for all object classes instead 
of separate codebooks for each class, described in more detail in Section \ref{sec:multiclass}.
 The gain of this is in terms of efficiency: Whereas the computational cost of separate detectors for each class grows 
linearly with the number of object classes, we can expect a sub-linear complexity for a multi-class detector. 
This is experimentally validated in Section \ref{sec:results_multiclass}.

Secondly, we include {\em depth constraints} in the voting for object centers. Depth measurements are obtained with 
a Kinect$\textsuperscript{\textregistered}$ camera. This increases the robustness of the detection, ruling out object 
hypotheses whose visual words are on significantly different depths. The underlying assumption is that the relative 
depths of different parts of an object are small in comparison to the distance to the camera. Furthermore, the encoded depth information also enables reasoning about object location in 3D, which is necessary for a humanoid robot application. This is further described in Section \ref{sec:depth}, and experimentally evaluated in Section \ref{sec:results_depth}.

\section{Related Work}
\label{sec:related}

The discussion of related work is divided into a review of detection work and one of segmentation. 

\vspace{1.7mm}
\noindent{\em Detection.~} 
Many object detection methods have been proposed in the past decades. 
The traditional approach is to use a sliding window over the whole image, accompanied by a binary classifier.
 Such an approach is very expensive and not scalable in terms of the number of categories. To improve efficiency, Viola and Jones 
\cite{Erdos06} use a boosted cascade of successively more elaborate classifiers. Most windows could be rejected in the early stages, while more complex, slower classifiers could focus on the few difficult cases. 
Extensions include, unsupervised online boosting \cite{Erdos17}  and kernel methods \cite{Erdos24}. 

A number of  bag-of-visual-words methods have also been proposed \cite{Erdos01,Erdos21,Erdos09,Erdos10}, where different ways  to learn a codebook including
 spatial information are proposed. There is in general a trade-off between structural flexibility in the model and the ability to capture structural information: A pure bag-of-words approach can not capture structural information, but is very robust to changes in shape, articulation and view point of objects. 
On the other hand, a method that models the metric relationships between visual words might capture spatial structure very well, but cannot generalize over changes in scale, viewpoint and articulation.
 
Leibe  et al.~\cite{Erdos01} (the basis of our method, described in Section \ref{sec:leibe}) found a good trade-off between flexibility and expressional power, using a voting based approach which enables detection of highly articulated objects in real-world
scenes. Object detection and segmentation are treated as two closely collaborating processes, improving each others' results. 
Developed in parallel with our approach,  Sun et al.~\cite{springerlink:10.1007/978-3-642-15555-0_48} present work that encodes depth information in this model, enabling shape recovery in 2D and 3D, and Razavi et al.~\cite{5995441} propose a scalable multi-class object detection by introducing a class dimension in the voting space. However, a robot detection system must be able to handle many object classes as well as reasoning about the objects in 3D. To that end, we present a humanoid household robot perception application (developed independently from \cite{springerlink:10.1007/978-3-642-15555-0_48,5995441}) which is multi-class scalable {\em and} encodes 3D information.      

Leibe  et al.~use a codebook which is built from textual information. Others \cite{Erdos21,Erdos27,Erdos23} build their visual words on contour-based information, with convincing
results. Opelt  et al.~\cite{Erdos21} present a boundary-fragment-model which uses a
contour-based features with information on the location of the object's centroid. 
By computing votes for the object centroid, boundary fragments are selected.
Shotton et al.~\cite{Erdos27} present a categorical object detection scheme that uses only local contour based features.  They employ a star-based configuration which is
flexible enough to cope with large variation in shape and appearance of both rigid and
articulated objects. Using the star model, of Felzenszwalb et al.~\cite{Erdos09,Erdos10} are able to detect rigid and deformable objects, using a topological grammar of object configuration. 

As discussed in Section \ref{sec:multiclass}, the codebook size is the key factor in the computational efficiency of a codebook-based detector. 
The codebook size can be decreased either by removing the least informative features \cite{Erdos36}, or by making sure that classes share features \cite{Erdos12}. Liu et al. \cite{Erdos54} present a general probabilistic framework for codebook selection which is shown to incorporate all entropy-based measures. 
In this paper 
we use information gain, which is defined in terms of the pointwise mutual information between words and object classes. A point for future work is to formulate this in terms of the general framework of \cite{Erdos54}.

Currently, using RGB-D imagery in Robotics draws a lot of attention. Lai et al.~\cite{Erdos55} recently published a RGB-D database and used visual cues, depth cues and rough knowledge of the configuration of the setting to segment objects in video sequences.  RGB and depth information are also used in {\em Instance Distance Learning} (IDL), proposed by Lai et al.~\cite{Erdos56}. It should be noted that IDL is only used for recognition, whereas our method addresses the more challenging problem of object detection in natural images. Furthermore, there are differences in the way depth features and visual features are integrated.

\vspace{2mm}
\noindent{\em Segmentation.~} 
It is argued that segmentation plays a fundamental role in human perception \cite{Erdos59} and is necessary for attention and detection tasks. 
Bottom-up segmentation,  where  image pixels are grouped, has been studied extensively in Computer Vision, but is by nature under-determined. We instead employ a top-down segmentation scheme guided by detection, where the method "knows what it is segmenting". 

Borenstein and Ullman \cite{Erdos61} propose a class-specific top-down segmentation which provides reliable results. They learn image fragments containing class and figure-ground information from training data, and then match these fragments to a test image. The patches together form a pixelwise foreground probability map which can be used to segment objects. 
Combining the top-down method \cite{Erdos61} with bottom-up segmentation, the method of Borenstein et al.~\cite{Erdos30} is able to combine the robustness of a top-down method with  the local detail of bottom-up segmentation.  

Bergstr{\"o}m and Kragic \cite{bergstrom:iros11} present an active 3D scene segmentation of unknown objects, as a basis for robot manipulation.
Using the assumption that objects are placed on flat surfaces, an image is segmented into three parts, object, surface and background.  
An extension \cite{Erdos67} segments several objects in a scene. 
However, one can argue that detecting the object class also is important for robot manipulation.

\begin{figure*}[t]
\centering
\subfigure{\includegraphics[width=0.244\textwidth]{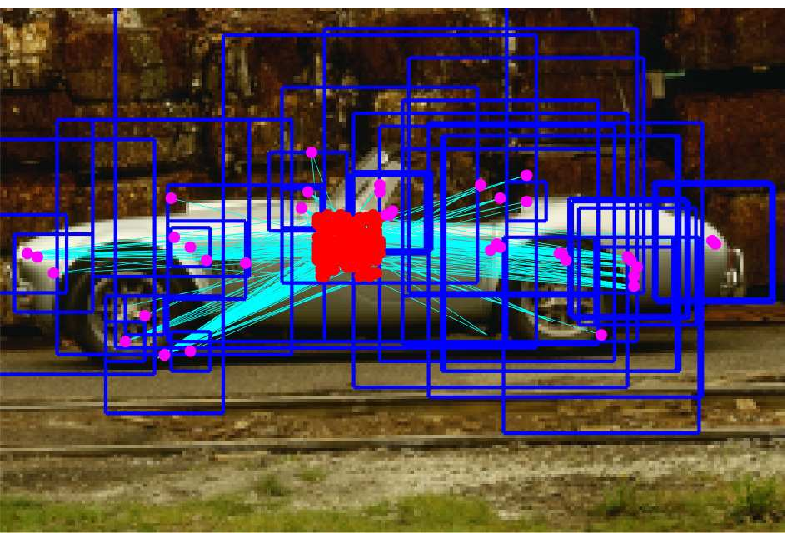}
\includegraphics[width=0.244\textwidth]{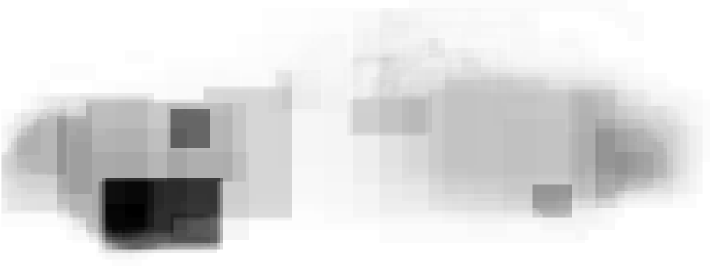} 
\includegraphics[width=0.244\textwidth]{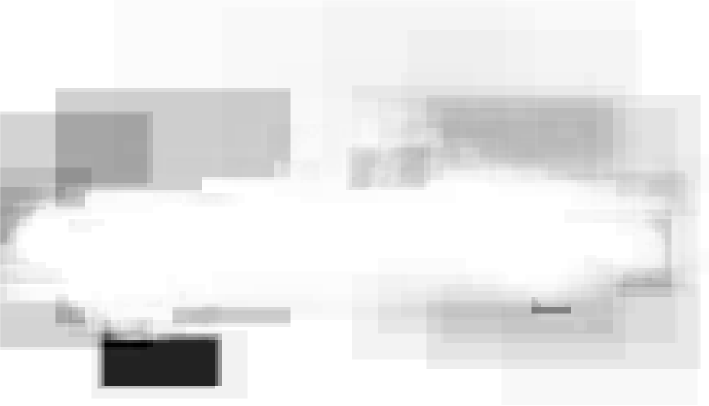}
\includegraphics[width=0.244\textwidth]{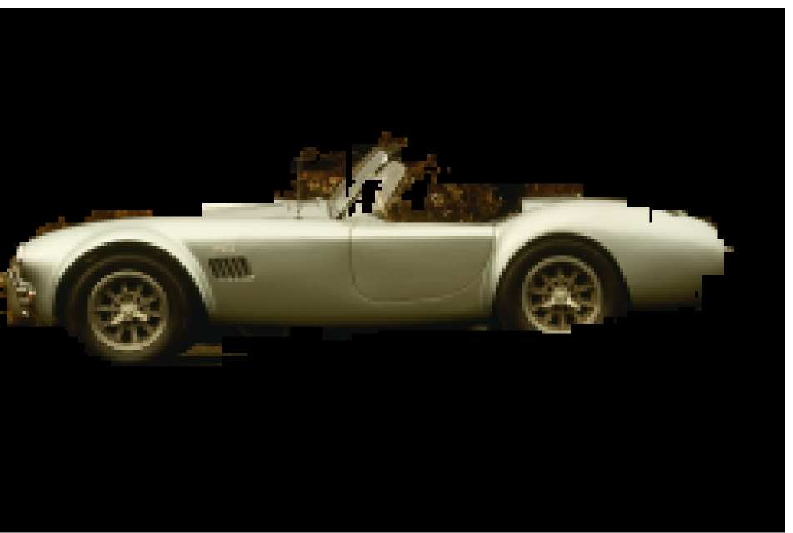}}
\caption{Detection example. The first image shows the matched features and voting back-projected from the hypothesis; the second image shows the object probability map; the third image shows the background probability map, computed on the matched feature areas. The last image shows the segmentation from the likelihood map.}
  \label{fig:segLeibe}
\end{figure*}

\section{Interleaved Segmentation and Detection}
\label{sec:leibe}

We base our work on the voting-based approach to object detection by interleaved segmentation and classification proposed by Leibe et al.~\cite{Erdos01}. The approach is briefly described here, see the reference for a more extensive description.


Objects are represented as a collection of visual words. 
The codebook of visual words for a certain object class is learned by clustering of SIFT  \cite{lowe04} descriptors extracted from training images of the object class. 

Based on the codebook, an {\em Implicit Shape Model} (ISM) is learned, encoding where on the object different features normally occur. For each visual word in the codebook, a distribution is learned over the spatial occurrence parameters
\begin{equation}
O^\mathit{ISM}_i = (x_i, y_i, s_i) \label{eq:o-s}
\end{equation}
where $(x_i, y_i)$ is the vertical and horizontal position relative to the object center and $s_i$ the SIFT feature scale.  Furthermore, each visual word is associated with a segmentation mask over the support area of the word feature.

When features are extracted from a new object image, each extracted feature is compared with the codebook entries. If the similarity to a word is above a certain threshold, it is allowed to vote for the most likely object center positions in the voting space 
\begin{equation}
V^\mathit{ISM}_i = (x_f - x_i\frac{s_f}{s_i}, y_f - y_i\frac{s_f}{s_i}, \frac{s_f}{s_i}) \label{eq:v-s}
\end{equation}
where $(x_f, y_f, s_f)$ are the position and scale of the extracted feature. The voting uses the learned distribution over $O^\mathit{ISM}_i$.
  
Mean Shift search is then applied to find the local maxima in the voting space, i.e., the most probable object centers. Figure \ref{fig:segLeibe} left shows a voting example.

The mutual confidence between the features and the object hypotheses is used to define a pixel-wise figure-ground segmentation. In a top-down manner, all features that contributed to the hypothesis are backprojected and their masks combined into foreground and background probability maps, as in Figure \ref{fig:segLeibe} center. A figure-ground segmentation, Figure \ref{fig:segLeibe} right, can be obtained from the ratio of these maps.

\section{Multi-Class Detection}
\label{sec:multiclass}

The ISM model of Leibe et al.~\cite{Erdos01} is designed to detect instances of a single class. To detect objects belonging to $n$ different classes, $n$ instances of
 the single class model can be trained with separate datasets of the different classes, and applied to the image independently of each other. 

However, the approach with separate detectors is not scalable, as the computation time is linearly dependent on the number of classes. A realistic robotic 
application would involve from 20 to several thousands of object categories -- rendering this approach is computationally infeasible. 

The approach that we propose here is to learn a joint model with a shared codebook and use the same voting space which contain a class dimention. Such a model can take advantage of the fact that similar features are present in different classes.


For large numbers of classes $n$, the codebooks of different individual class detectors are likely contain very similar features, corresponding to often-occurring patterns such as straight lines, corners, etc. With a joint codebook, these features could be shared between different classes, increasing computational efficiency; the codebook size is the key issue for computation cost since the detection is based on matching codebook entries. The main advantage with a model with shared codebook is thus the decrease in computational complexity, since more and more features are shared as $n$ grows.  

Interest points are extracted from images of all $n$ classes, as in the single class method. The codebook is then learned from clustering all features together using RNN clustering. 

\begin{figure*}[t]
\centering
\subfigure[Single class detectors, accuracy 85\%, $T_Q$=0, 7825 words]
{\includegraphics[height=24mm]{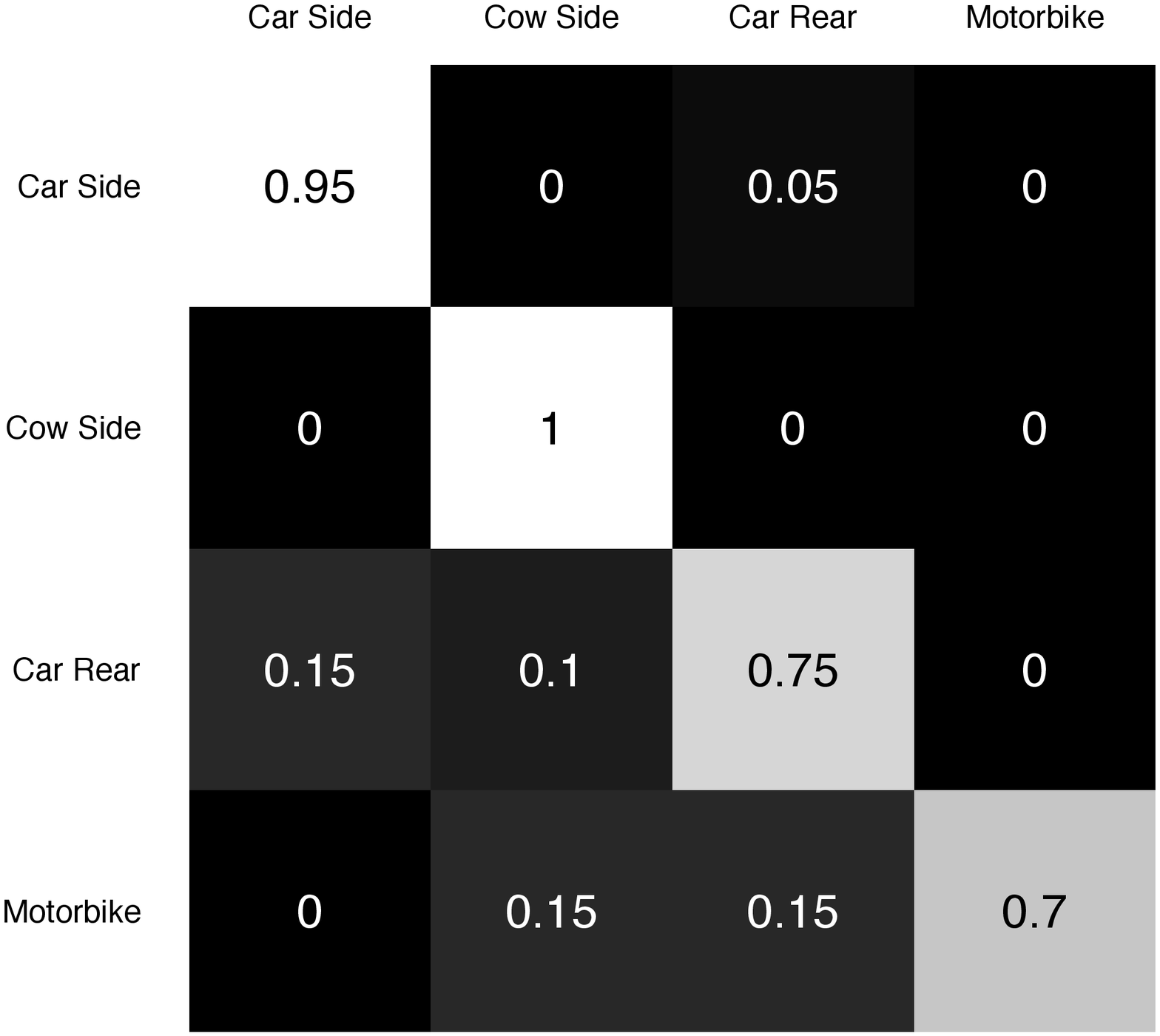}
 \label{fig:cf_sep_0} }
\subfigure[Multi-class detector, accuracy 81\%, $T_Q$=0, 6949 words]
{\includegraphics[height=24mm]{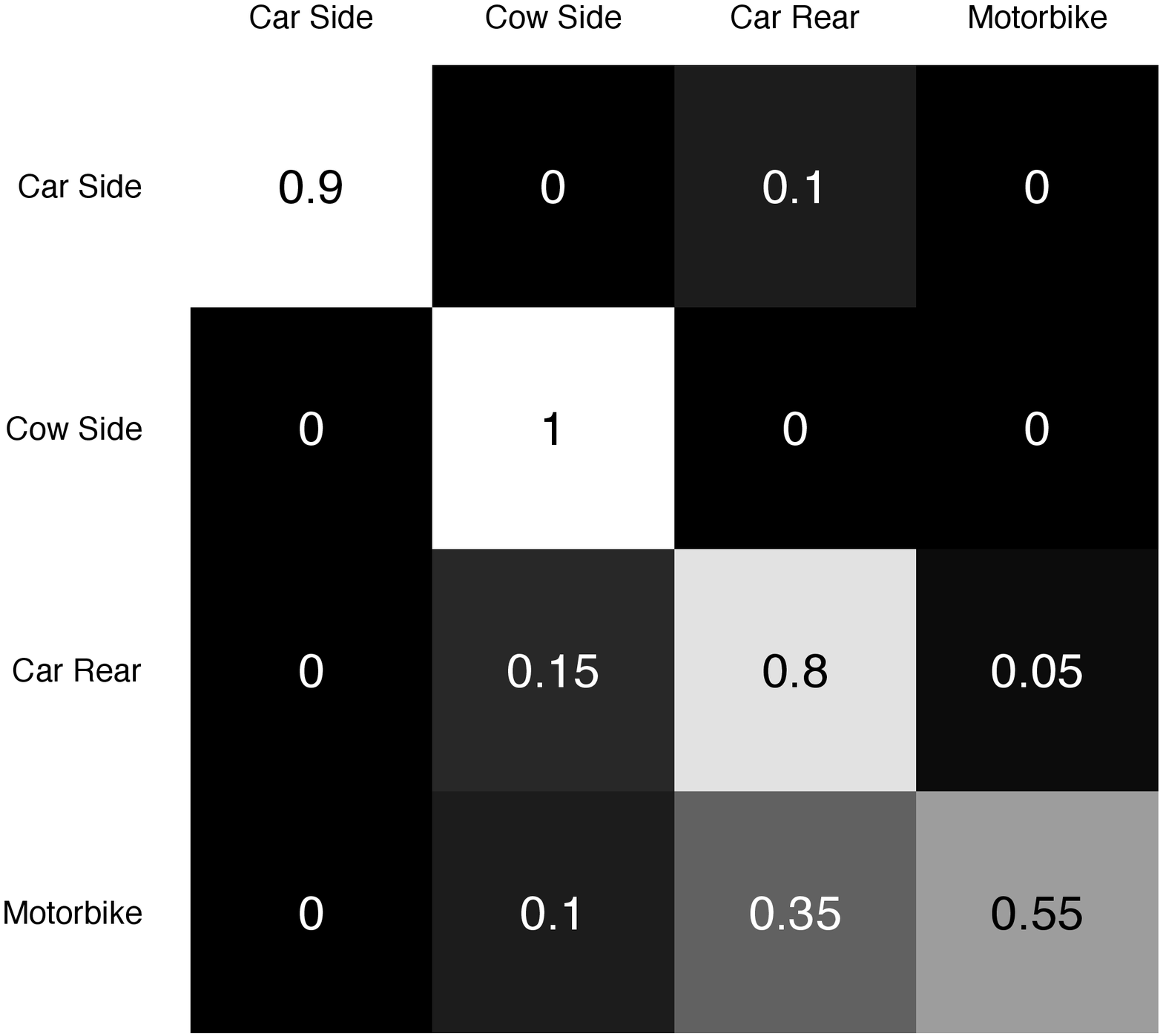}
\label{fig:cf_mul_0}}
\subfigure[Single class detectors, accuracy 81\%, $T_Q$=1, 4777 words]
{\includegraphics[height=24mm]{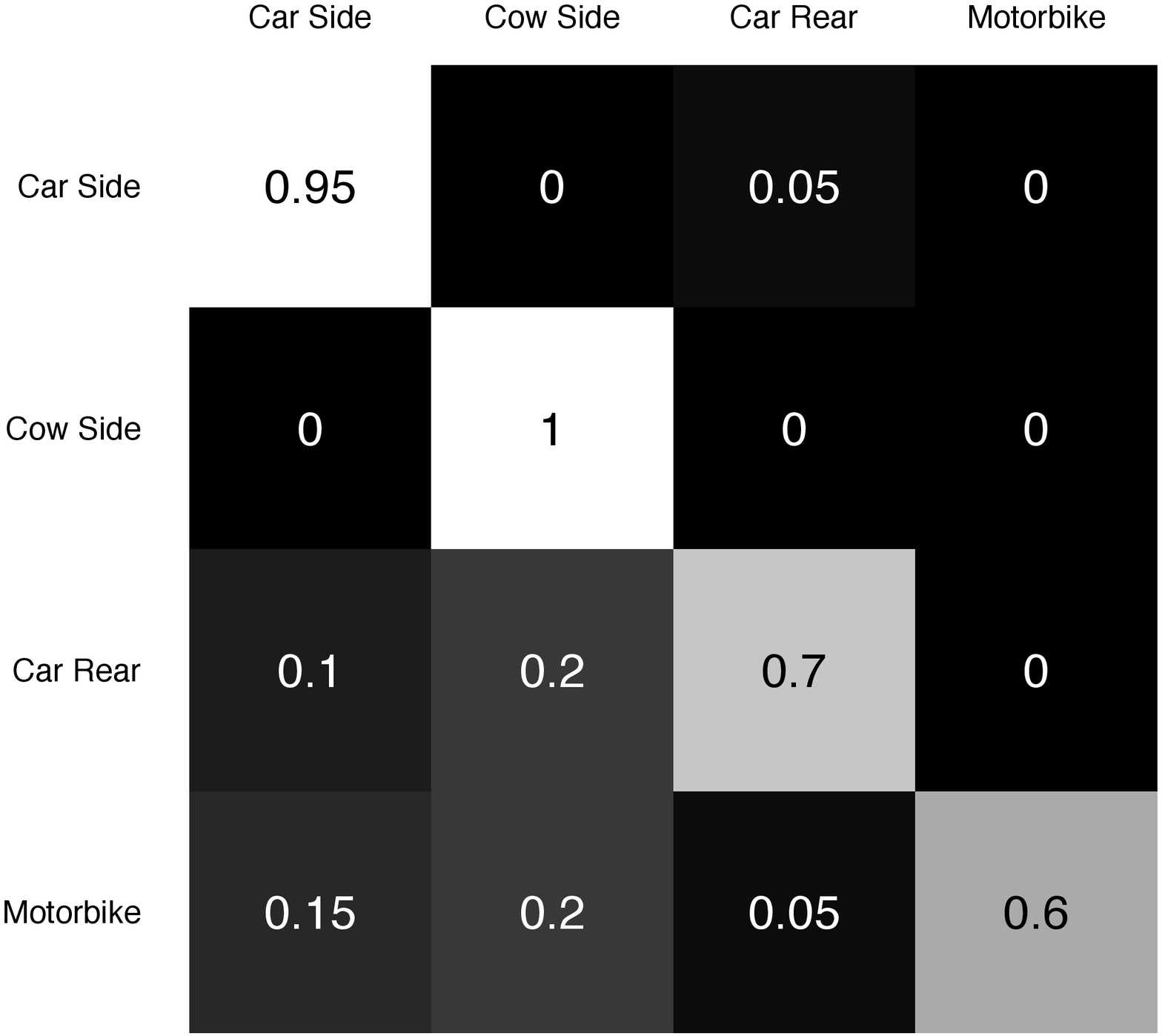}
 \label{fig:cf_sep_1} }
\subfigure[Multi-class detector, accuracy 88\%, $T_Q$=1, 4578 words]
{\includegraphics[height=24mm]{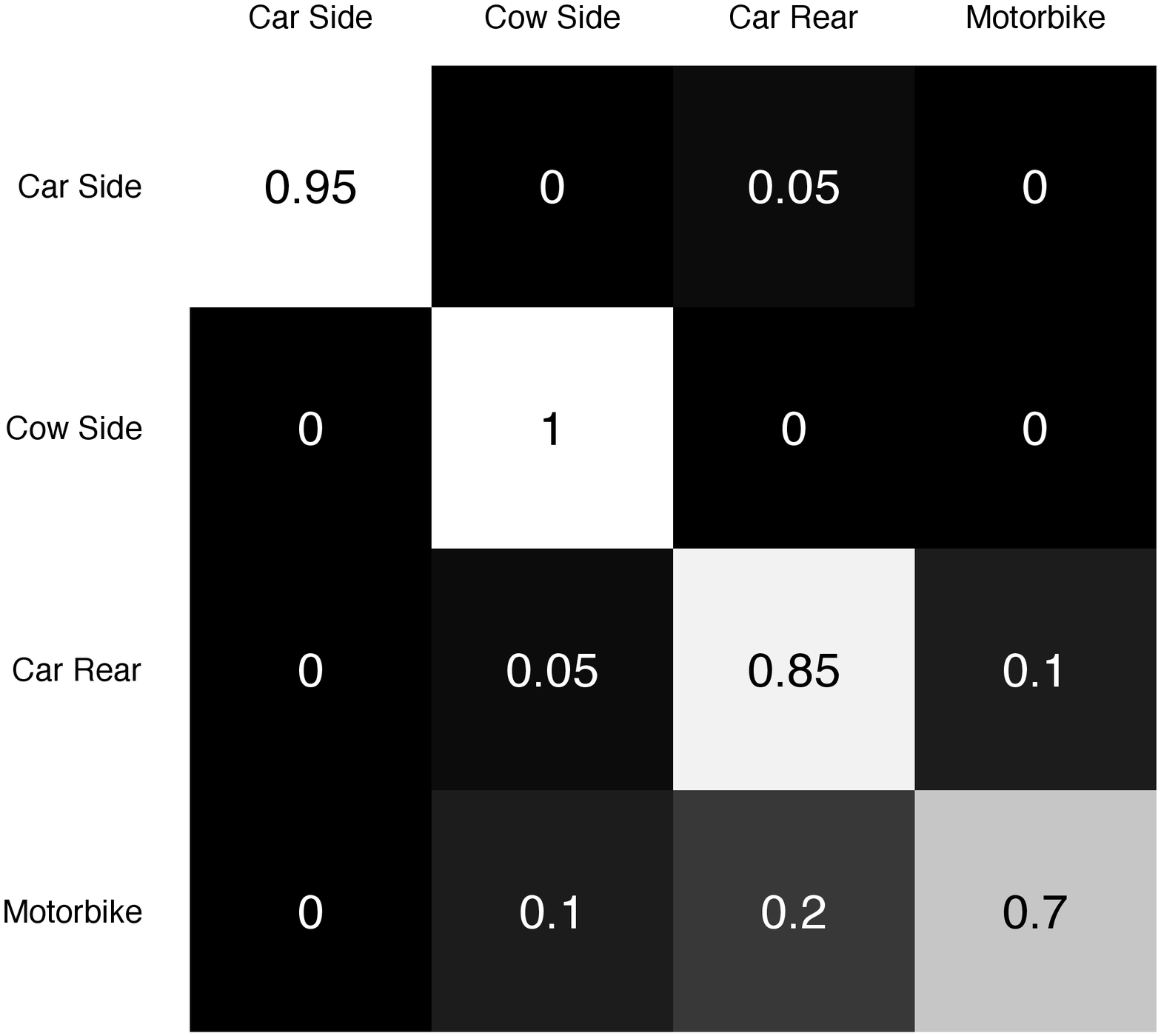}
\label{fig:cf_mul_1}}
\subfigure[Single class detectors, accuracy 83\%, $T_Q$=2, 2748 words]
{\includegraphics[height=24mm]{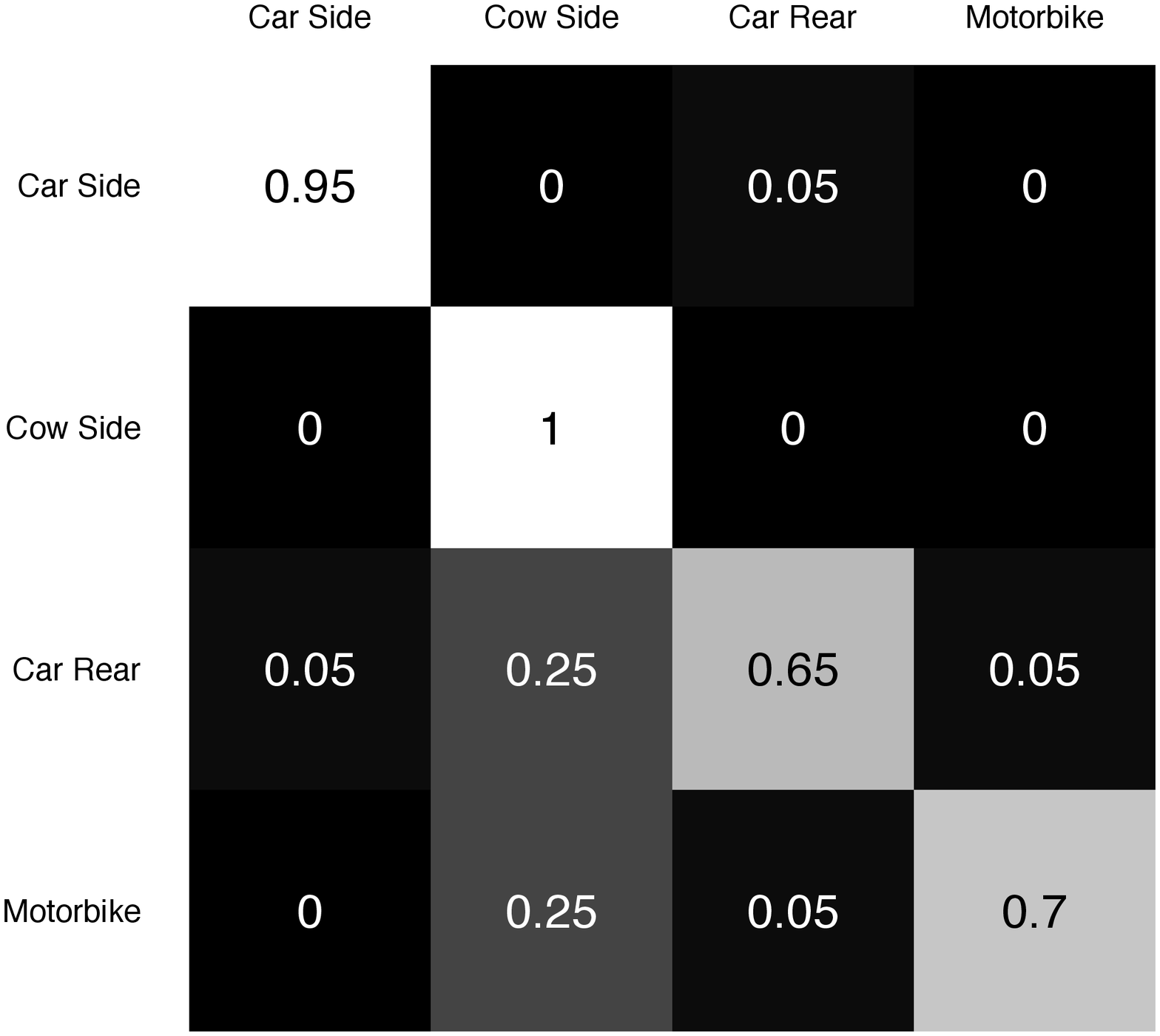}
 \label{fig:cf_sep_2} }
\subfigure[Multi-class detector, accuracy 83\%, $T_Q$=2, 2680 words]
{\includegraphics[height=24mm]{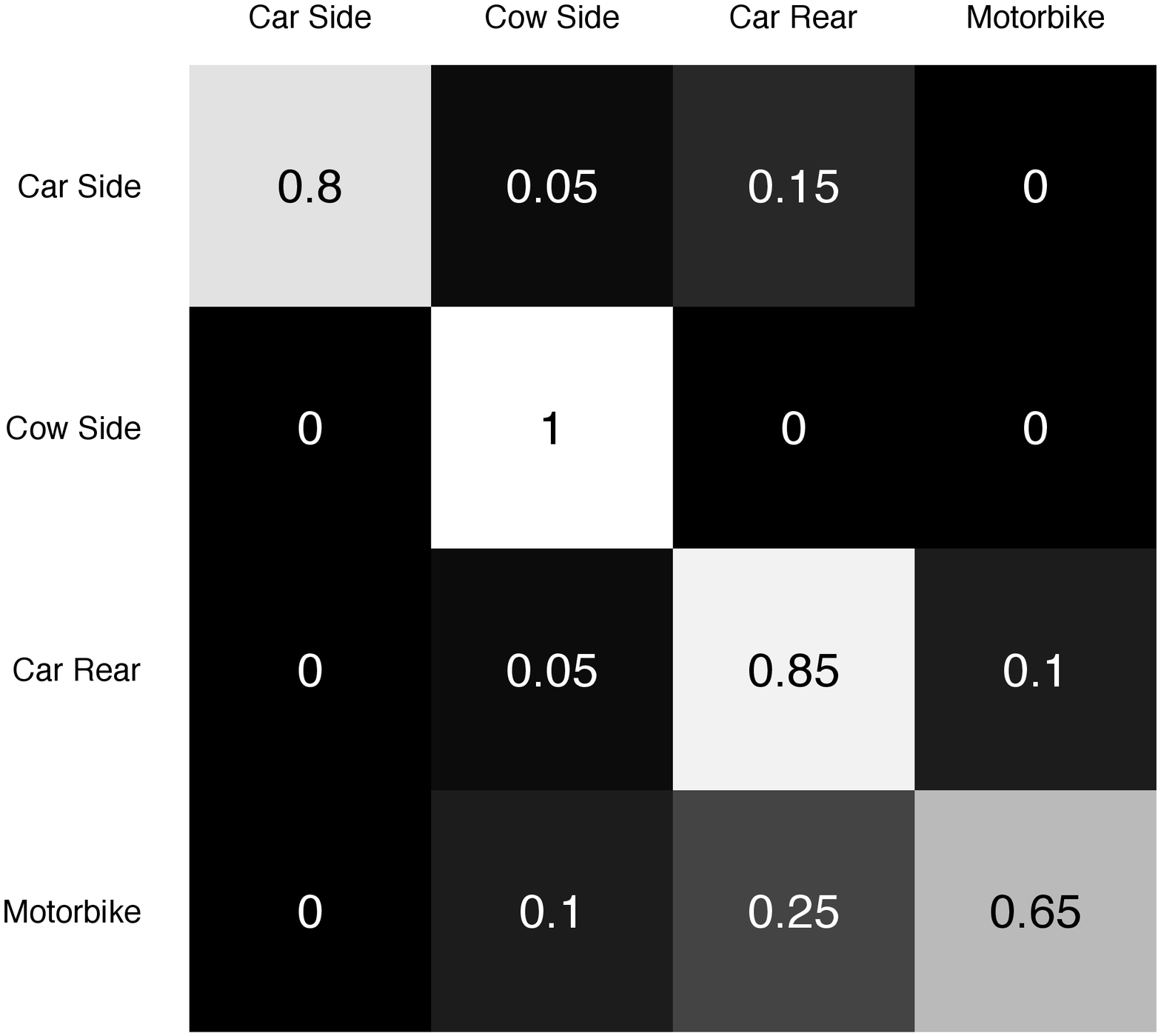}
\label{fig:cf_mul_2}}
\caption{Classification performance, confusion matrices for different parameter
settings.  $T_Q$ is the threshold on codebook cluster size.}
\label{fig:confusion}
\end{figure*}

As we shall see in Figure \ref{fig:confusion}, the gain in codebook size (i.e., the ratio of features shared between different classes) 
is quite moderate when performing "raw"  RNN on the joint feature set. However, the codebook size can be further decreased. Given the thesis that many 
features are in fact shared among classes, a reasonable assumption is that many of the words in the codebook are present on instances of many of the $n$ classes -- and also for other types 
of objects, or in the background of images. Hence, they are not very descriptive of a certain 
class, or not even of the foreground areas in general. The detection performance will then not be affected if these words are removed.

The principled approach to removing uninformative codebook entries is to measure the mutual information between words and class labels \cite{Erdos36,Erdos37}. The {\em information gain} $G$ of a certain word $w_i$ is equal to the average pointwise mutual information between the word and all class labels,
\begin{equation}
 G(w_i)= \frac{1}{n}\sum_{c=1}^n  P(w_i, c) \log \frac{P(w_i, c)}{P(w_i) P(c)} ~.
 \end{equation}
Words with a high gain $G(w_i)$ are specific to a certain class, and thus correspond to unusual patterns with high discriminative power. 

A principled way of decreasing codebook size is thus to remove words with an information gain lower than a certain threshold. The effects on performance of different information gain thresholds are evaluated in Section \ref{sec:results_multiclass}.

Based on the joint codebook, a {\em Joint Implicit Shape Model} (JISM) is learned, encoding on which objects, and where on the object different features are likely occur. A class parameter is added to the spatial occurrence distribution of codebook entries; for each visual word in the joint codebook, a distribution is learned over the spatial occurrences
\begin{equation}
O^\mathit{JISM}_i = (x_i, y_i, s_i, c_i) \label{eq:o_sandc}
\end{equation}
where $(x_i, y_i)$ is the vertical and horizontal position relative to the object center, $s_i$ the SIFT feature scale, and $c_i$ the class of the object.


As in the original method in Section \ref{sec:leibe}, the first step in object detection is to extract features in the image, and match these to the joint codebook. 
Using the learned distributions over $O^\mathit{JISM}_i $, the features vote for object center positions, scales and classes in the voting space
\begin{equation}
V^\mathit{JISM}_i = (x_f - x_i\frac{s_f}{s_i}, y_f - y_i\frac{s_f}{s_i}, \frac{s_f}{s_i}, c_i) \label{eq:v-sandc}
\end{equation}
in the same manner as in the original method. 

Moreover, only patches belonging to the object class corresponding to the majority vote are employed for segmentation -- others are considered as false positives. 

\section{3D Detection}
\label{sec:depth}

The JISM model described above suffers from different forms of noise as ISM, e.g., spurious feature detections in the background, or erroneous association of features from two different objects (for example, see Figure \ref{fig:2D_vs_sd}(a,d), where the space between the front wheels of the left car and the rear wheels of the right car is wrongly detected as a car). 

This is addressed by using depth. Provided measurements of depth for every extracted feature, e.g., from RGB-D imagery captured with a Kinect$\textsuperscript{\textregistered}$ sensor, depth can be included among the spatial occurrence and class parameters, giving us an {\em Joint Implicit 3D Shape Model} (JI3SM). 

The assumption underlying the JI3SM is that only features that are on similar depth can vote for the same object position. This means that depth is only explicitly involved in the detection phase, not in training. At training time, distributions over $O^\mathit{JI3SM}_i = O^\mathit{JISM}_i = (x_i, y_i, s_i,c_i)$ are learned as above. 

(It would be possible to relax the assumption about similar feature depth by learning feature depths $d_i$ relative to the object center, and add these to the measured absolute feature depth $d_f$ in the detection stage. This would require a depth map associated with each training object instance.)


As described above, features are extracted and matched to the codebook.  However, every feature is assigned a depth marking the feature location in 3D space.

The depth and scale parameters of features are (inversely) correlated, but contain slightly different information. In the light of this, we suggest the following three alternatives for augmenting the voting space.
\begin{eqnarray}
&&\hspace{-1.2cm}\mathrm{Alt~1:} ~~~ V^\mathit{JI3SM\,1}_i = (x_f - x_i\frac{s_f}{s_i}, y_f - y_i\frac{s_f}{s_i}, \frac{s_f}{s_i}, d_f, c_i) \label{eq:v-sandd}\\
&&\hspace{-1.2cm}\mathrm{Alt~2:} ~~~ V^\mathit{JI3SM\,2}_i = (x_f - x_i\frac{s_f}{s_i}, y_f - y_i\frac{s_f}{s_i}, d_f, c_i) \label{eq:v-d}\\
&&\hspace{-1.2cm}\mathrm{Alt~3:} ~~~ V^\mathit{JI3SM\,3}_i = (x_f - x_i\frac{s_f}{s_i}, y_f - y_i\frac{s_f}{s_i}, d_f\frac{s_f}{s_i}, c_i) \label{eq:v-stimesd}
\end{eqnarray}
where $(x_f, y_f, s_f, d_f)$ are the position, scale and absolute distance to the camera of the extracted feature. 

The depth cue is expected to increase the robustness and accuracy of detection, since incorrect object hypotheses are less likely to appear; spurious features that accidentally support an incorrect object hypothesis rarely lie on the same depth range, On the other hand, correct features lie in the same depth range (given that the similar depth assumption holds), and are allowed to support each others votes.

The segmentation step does not currently involve depth information. A focus of future research is to explore depth information for segmentation; there is of course a high correlation between object boundaries and depth boundaries. Depth boundaries are characterized by empty areas (see Figure \ref{fig:sd3}), which create easily detectable "halos" around objects. This is further discussed in the Conclusions.

\begin{figure*}[t]
\centering
\subfigure{\includegraphics[height=9.5mm]{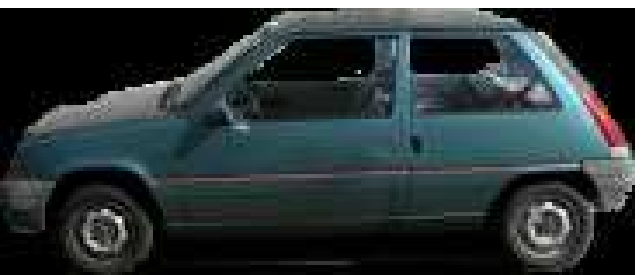}
\includegraphics[height=9.5mm]{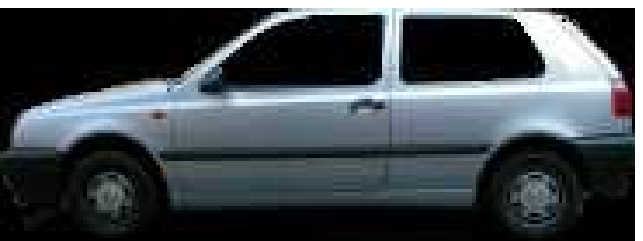}
\includegraphics[height=9.5mm]{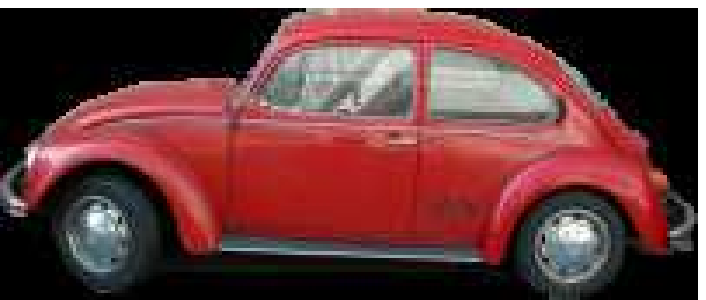}
\includegraphics[height=9.5mm]{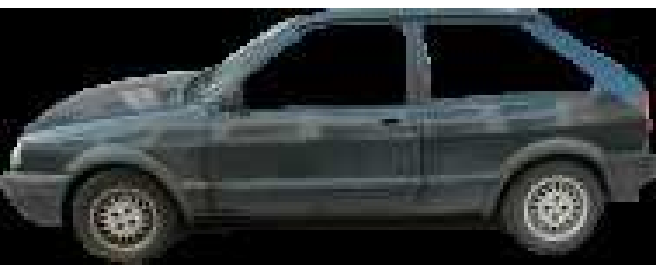}
\includegraphics[height=9.5mm]{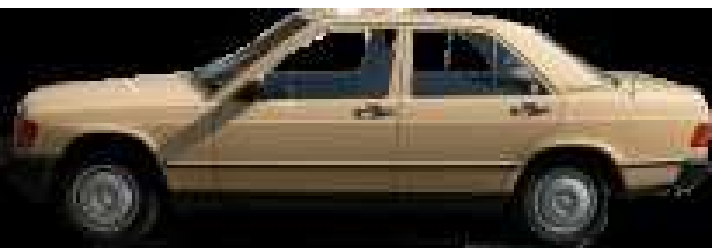}
\includegraphics[height=9.5mm]{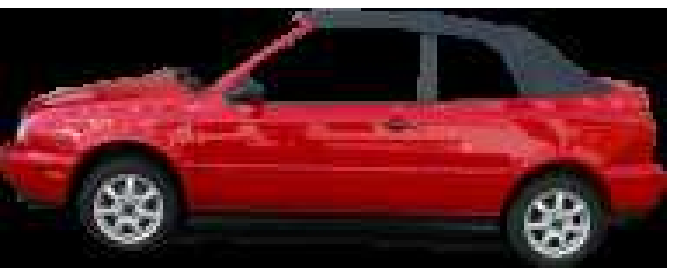}
\includegraphics[height=9.5mm]{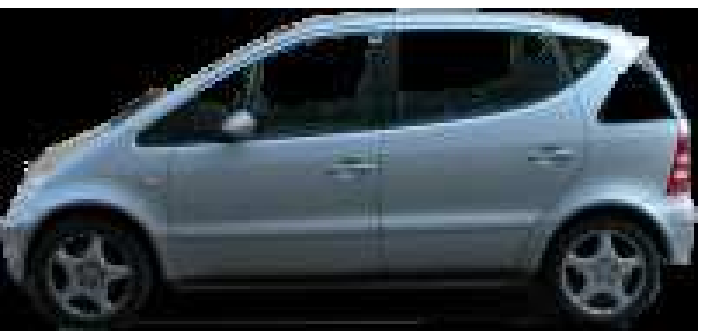}}
\subfigure{\includegraphics[height=10.45mm]{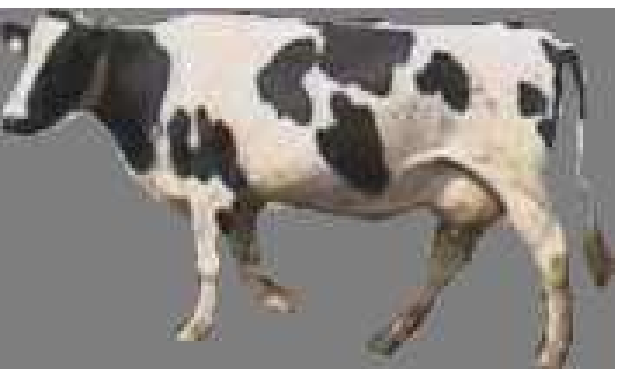}
\includegraphics[height=10.45mm]{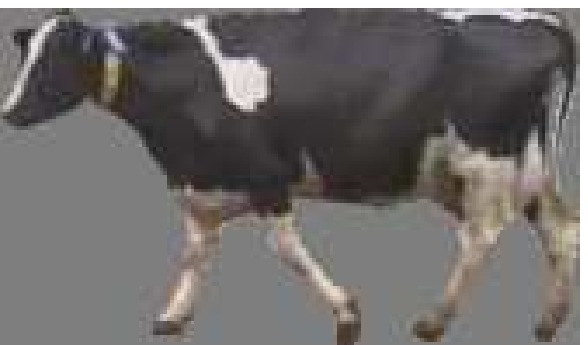}
\includegraphics[height=10.45mm]{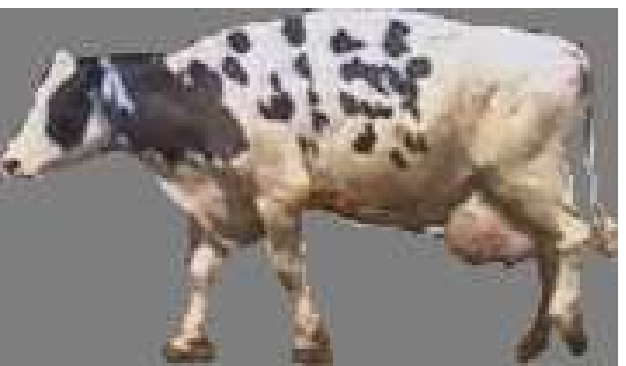}
\includegraphics[height=10.45mm]{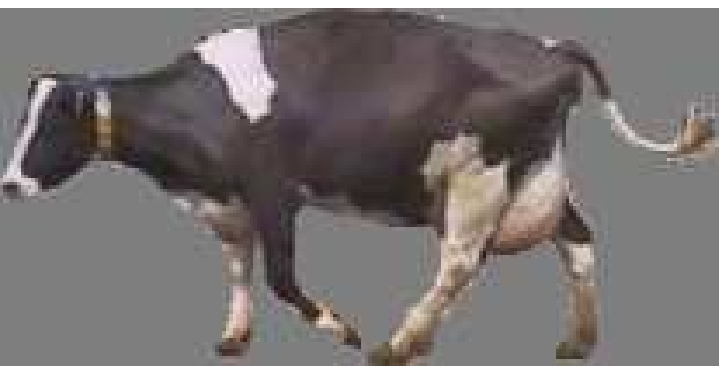}
\includegraphics[height=10.45mm]{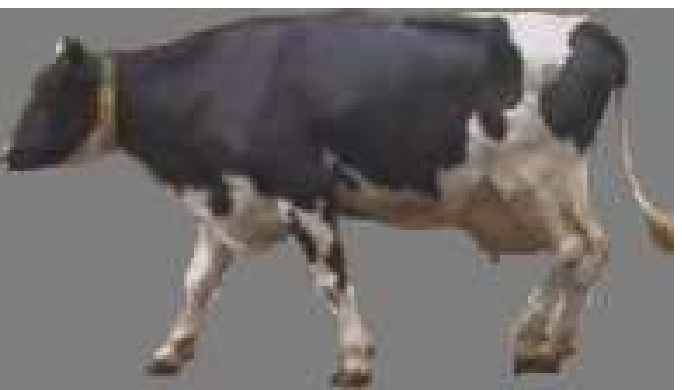}
\includegraphics[height=10.45mm]{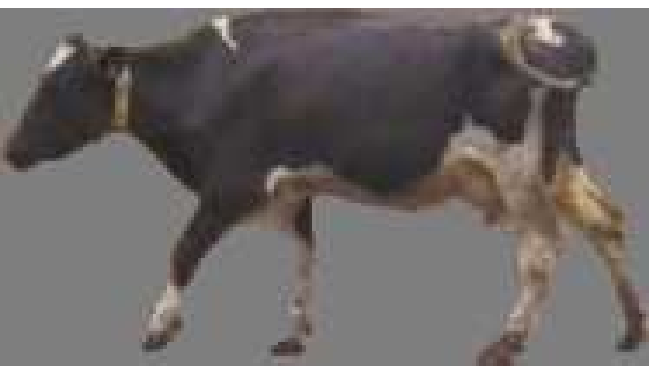}
\includegraphics[height=10.45mm]{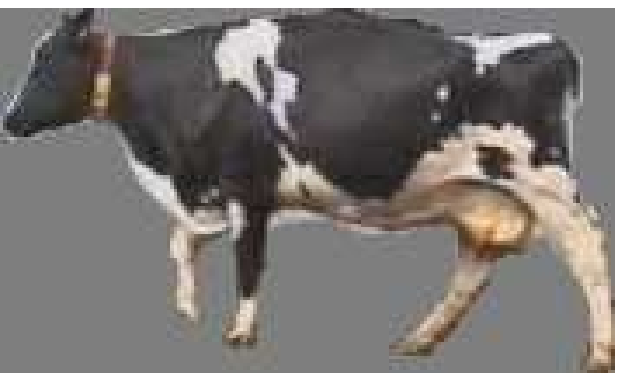}
\includegraphics[height=10.45mm]{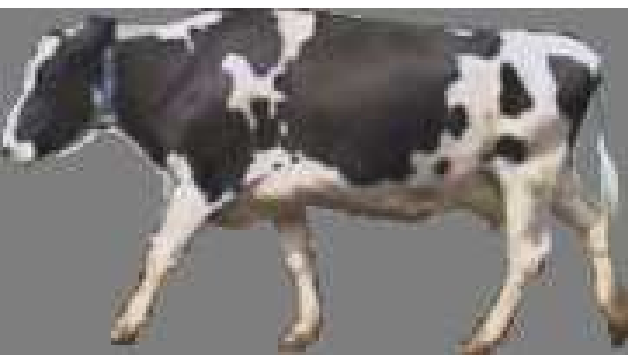}
\includegraphics[height=10.45mm]{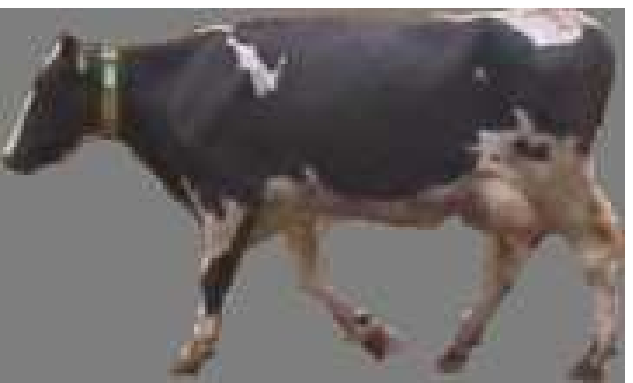}}
\subfigure{\includegraphics[height=10.99mm]{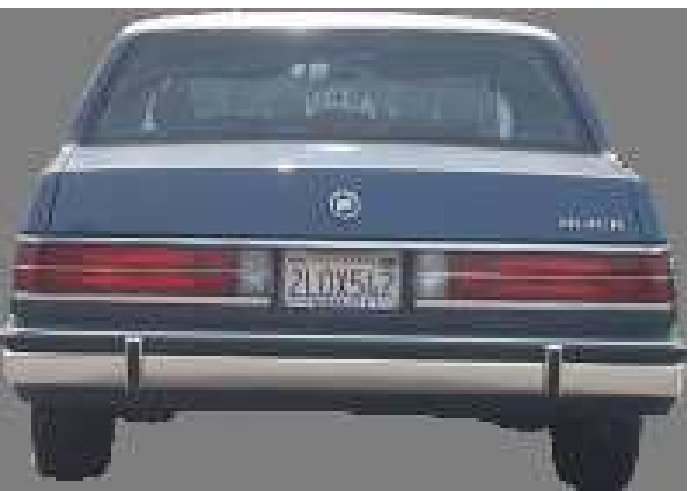}
\includegraphics[height=10.99mm]{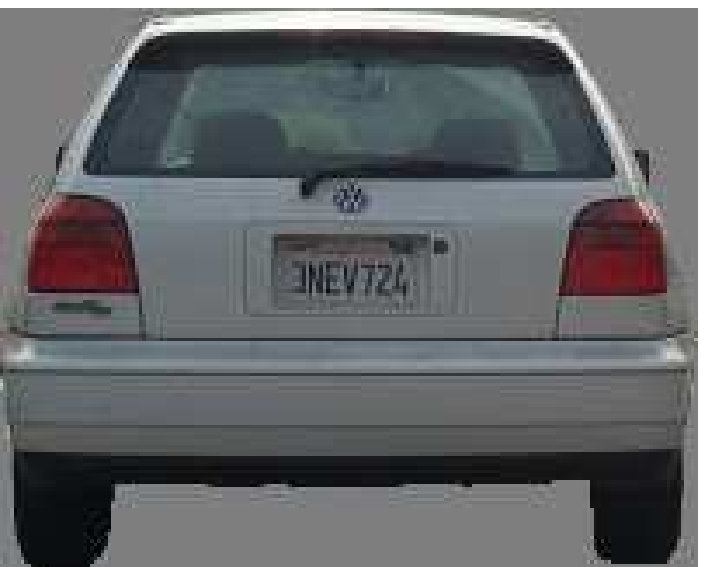}
\includegraphics[height=10.99mm]{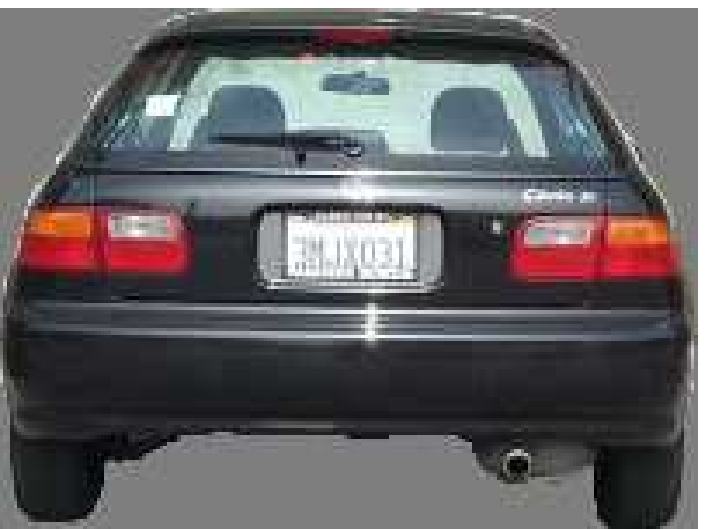}
\includegraphics[height=10.99mm]{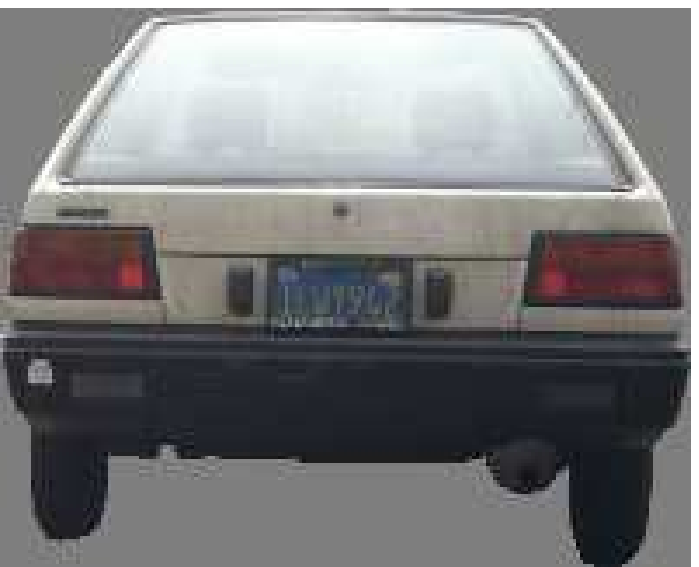}
\includegraphics[height=10.99mm]{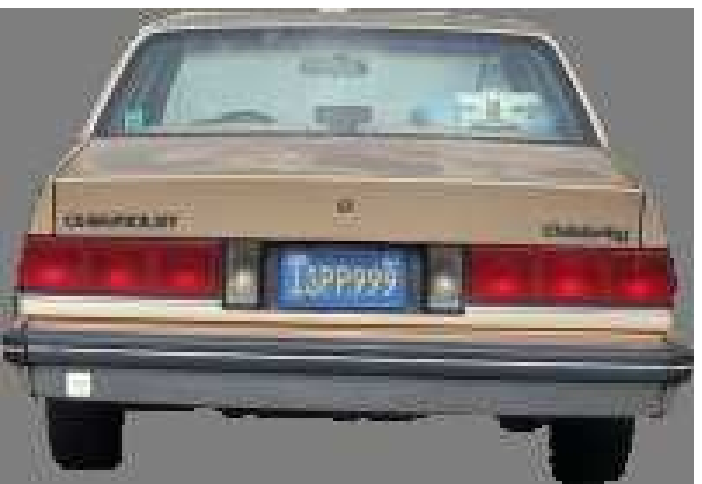}
\includegraphics[height=10.99mm]{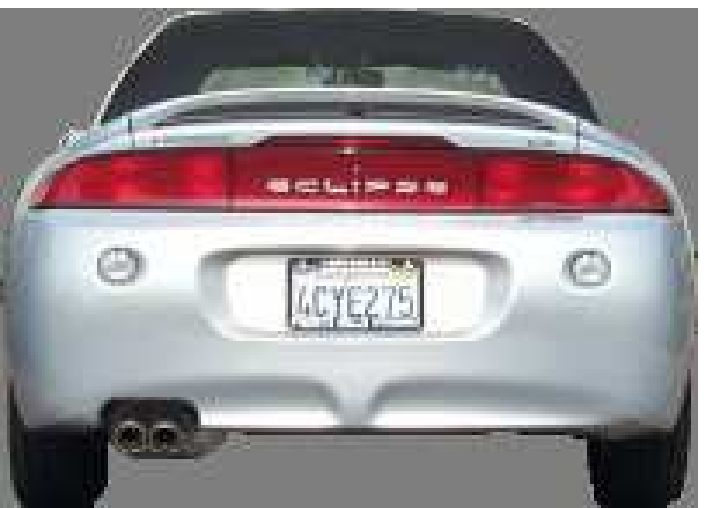}
\includegraphics[height=10.99mm]{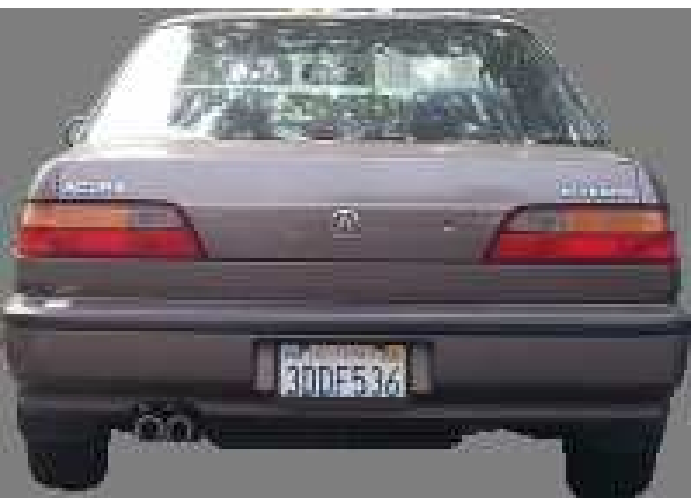}
\includegraphics[height=10.99mm]{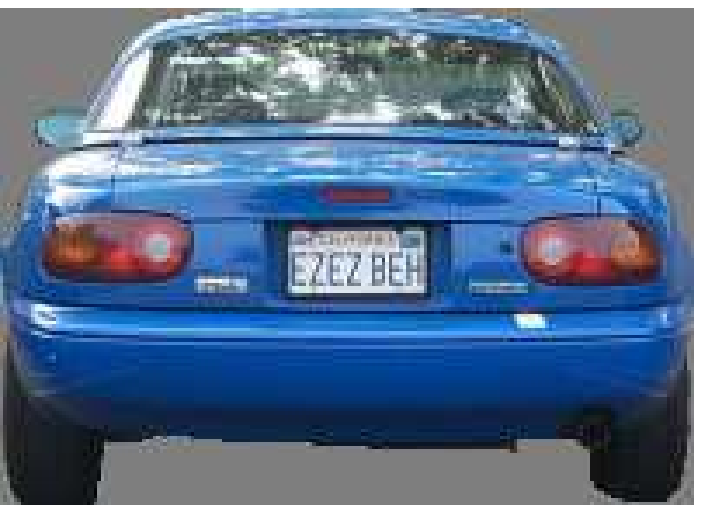}
\includegraphics[height=10.99mm]{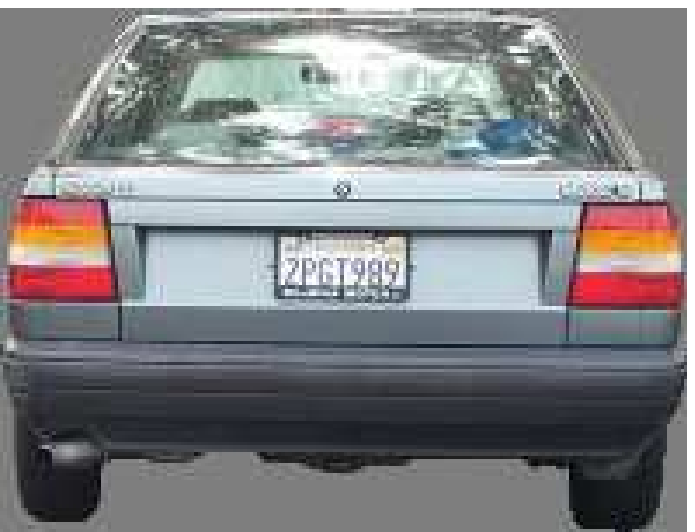}
\includegraphics[height=10.99mm]{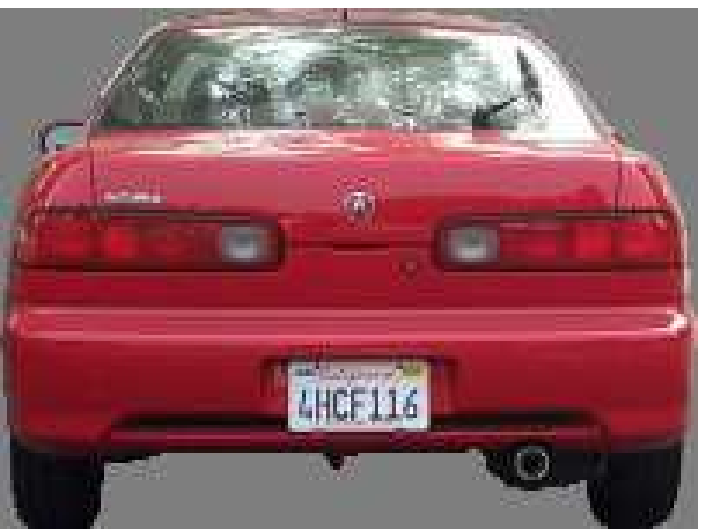}
\includegraphics[height=10.99mm]{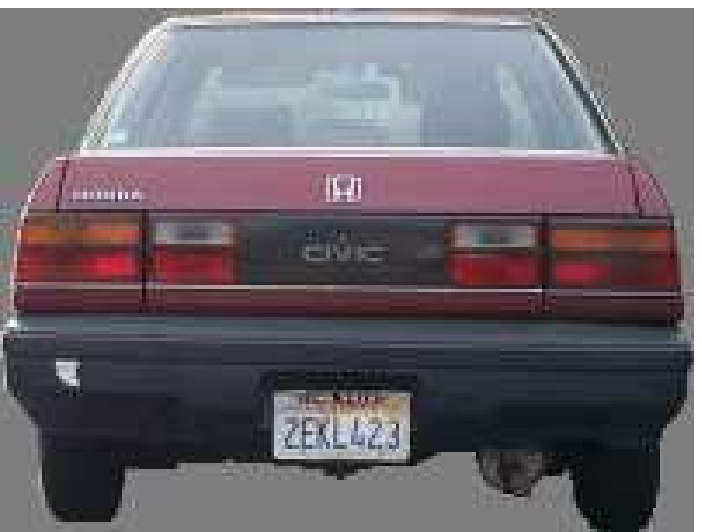}}
\subfigure{\includegraphics[height=10.195mm]{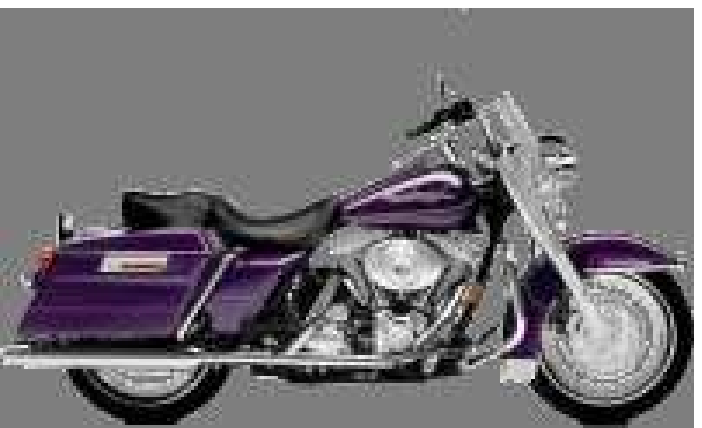}
\includegraphics[height=10.195mm]{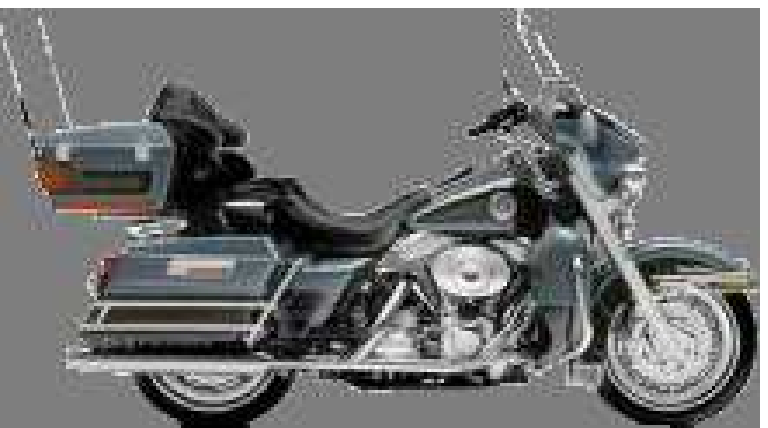}
\includegraphics[height=10.195mm]{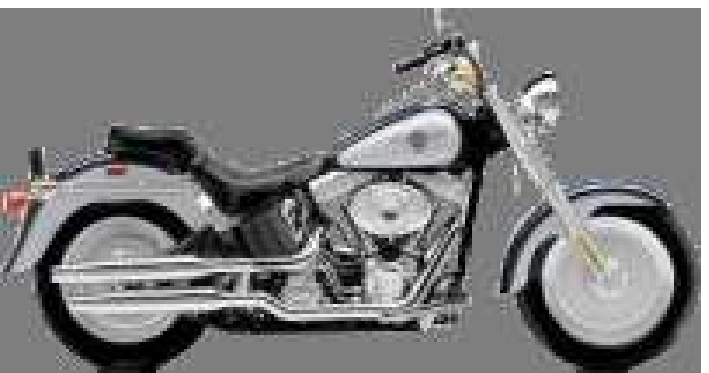}
\includegraphics[height=10.195mm]{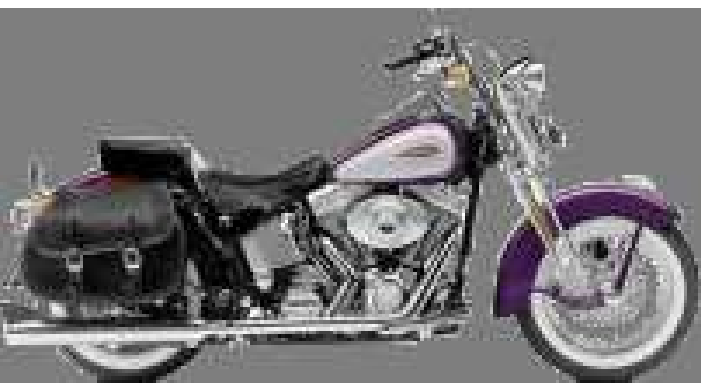}
\includegraphics[height=10.195mm]{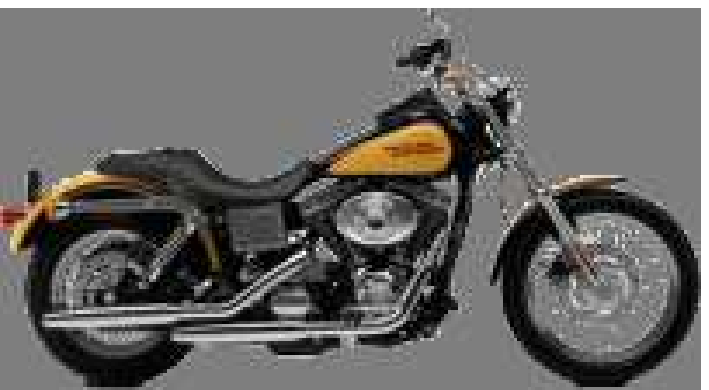}
\includegraphics[height=10.195mm]{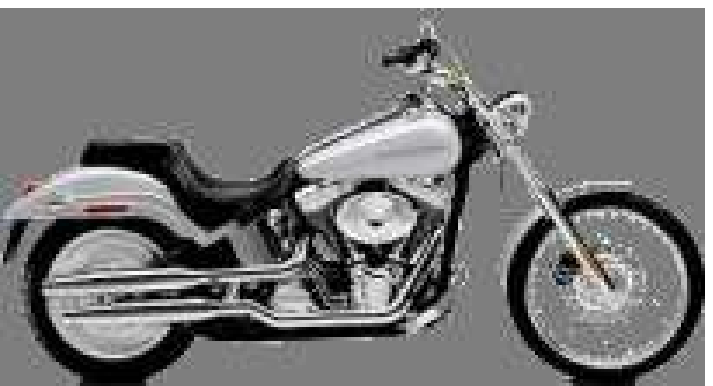}
\includegraphics[height=10.195mm]{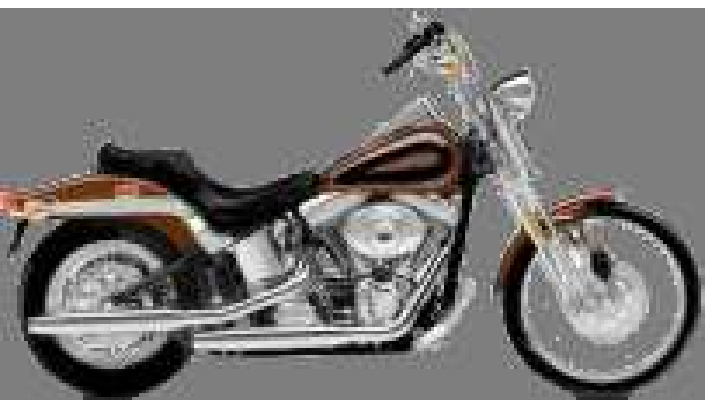}
\includegraphics[height=10.195mm]{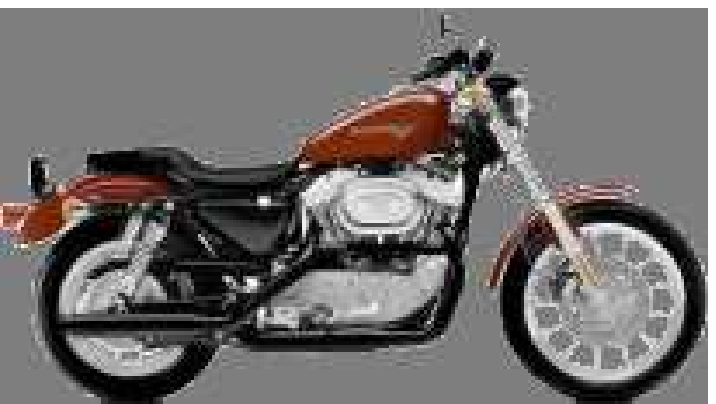}
\includegraphics[height=10.195mm]{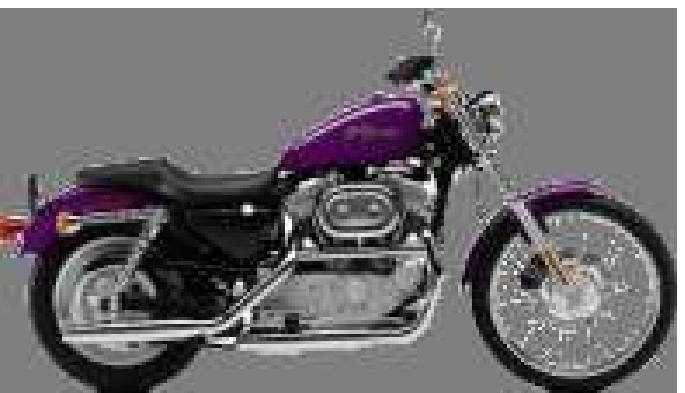}}
\caption{Examples from the multi-class training dataset [1].}
\label{fig:LeibeExample}
\end{figure*}

\section{Results}
\label{sec:results}

The JISM and JI3SM were implemented in C++ on a regular desktop machine.

\subsection{Multi-Class Detection}
\label{sec:results_multiclass}

\begin{figure}[t]
\centering
\includegraphics[width=\linewidth]{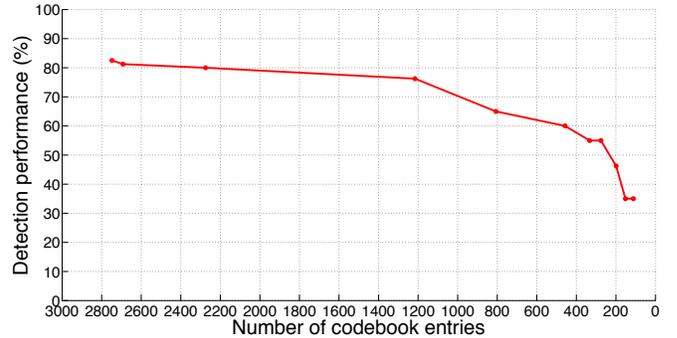}
\caption{Classification performance, mean classification accuracy for different
parameter settings. $T_\mathit{IG}$ is the threshold on information gain, and
varies (from left to right) from 0 to 0.01. $T_Q$=2.}
\label{fig:ig_graph}
\end{figure}

The JISM described in Section \ref{sec:multiclass} was evaluated using a
four-class dataset of objects in natural outdoor settings. No depth was
available due to the limitation of the Kinect$\textsuperscript{\textregistered}$
sensor to indoor scenes.

The training set consisted of 30 images of each of the classes {\tt car side},
{\tt cow side}, {\tt car rear} and {\tt motorbike}, and sampled from the dataset
 {\tt http://www.vision.ee.ethz.ch/$\sim$bleibe/data /datasets.html}, see Figure
\ref{fig:LeibeExample}. All images were labeled with a segmentation mask. For
testing,  20 unseen images (not in the training set) of the three first classes 
were chosen randomly from the same dataset. To increase the difficulty of the
{\tt motorbike} test cases,  20 single motorbikes images were chosen randomly
from the TUD dataset {\tt http://www.mis.tu-darmstadt.de/datasets}. 

The JISM was trained using the training set above. For comparison, four
individual ISM instances were also trained with each of the four classes of the
training set. The Harris-Laplace detector and SIFT descriptor from
\cite{Erdos40} were used for feature extraction.

In this experiment, only the classification aspect of the detection was
evaluated. All four ISMs were applied to each image in the test set, and the
object hypothesis 
As exawith highest score was regarded as the classification result of that
image. Figure \ref{fig:cf_sep_0} shows the classification result with
the multiple single class models and Figure \ref{fig:cf_mul_0} shows the
classification result with the multi-class model.
We can see that the same level of detection accuracy can be achieved with
smaller codebook size.
The single class model codebooks contain 7825 entries together, while the
multi-class codebook size is only 6949. The reason for the 10\% decrease in
codebook size is 
that words are shared between classes. It should be noted that the number of
classes is quite small -- more classes should lead to a larger decrease.
Further, we will discuss that 
with multi-class detection, many existing codebook selection method can be
easily applied.

\begin{figure*}[t]
\centering
\includegraphics[width=160mm]{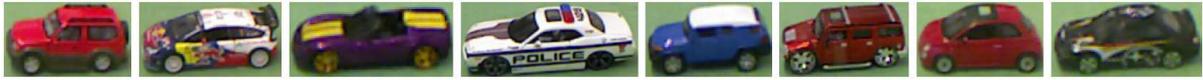} 
\caption{Examples from the 3D toy car dataset.}
\label{fig:3DTrainingExample}
\end{figure*}

\begin{figure*}[t]
\centering
\includegraphics[width=160mm]{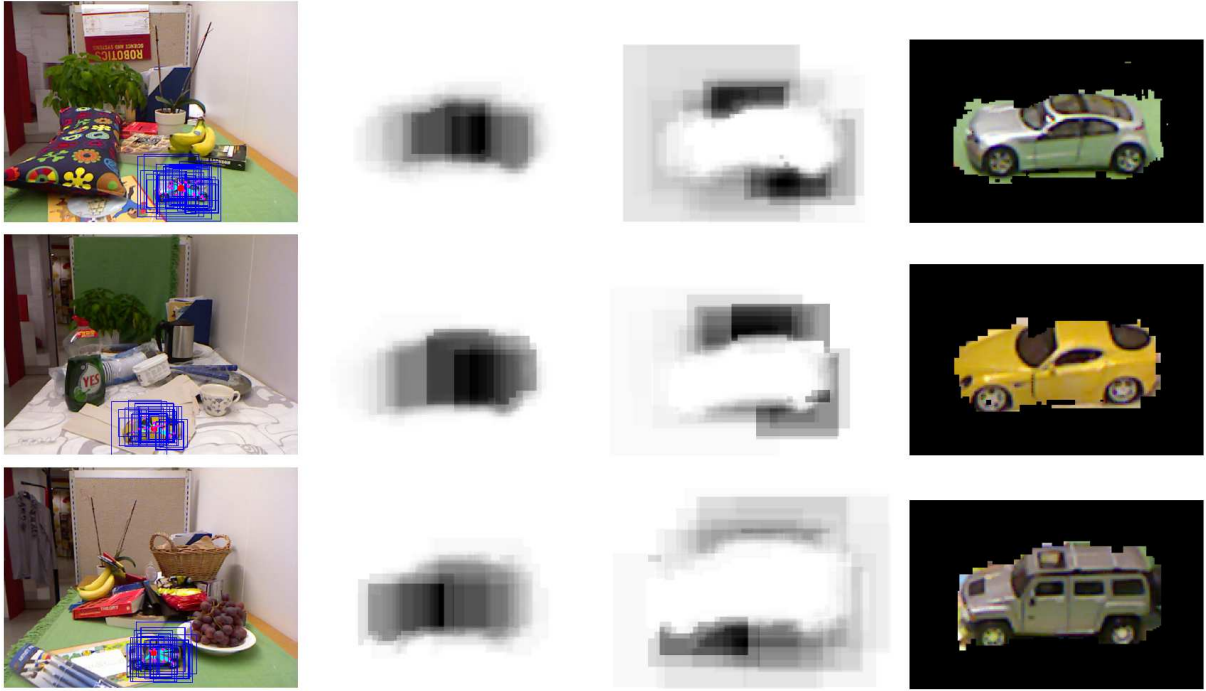} 
\caption{Detection examples using $V^\mathit{JI3SM\,3}$ (Eq (8)). For every row, the first image shows the matched features and voting back projected from the hypothesis; the second image shows the object probability map; the third image shows the background probability map, computed on the matched feature areas; the last image shows the segmentation from the likelihood map.  To lower the number of detections and the size of the detection for illustrative purposes, a smaller bandwidth was used than in Table 1: 5 for $x$ and $y$, and 0.07 for $s\times d$.}
  \label{fig:segEg}
\end{figure*}

It is argued \cite{Erdos01} that it is safe to ignore clusters with only one feature, as they likely correspond to class outliers.  Figure \ref{fig:cf_sep_1}  shows the detection result with
the multiple single class models and codebooks with clusters containing $>1$ feature, and Figure \ref{fig:cf_mul_1}  shows the detection result with the multi-class model and the joint codebook with clusters containing $>1$ feature. The multi-class model performed slightly better than the single class model. However, the decrease in codebook size was smaller with thresholding, since one-feature clusters are less likely with a larger set of features to cluster -- removing clusters with 1 feature in the single class case might be comparable to removing multi-class clusters with  $\leq2$ features. Figures \ref{fig:cf_sep_2} and \ref{fig:cf_mul_2} show the result of removing clusters of size $\leq2$. 

As discussed in Section \ref{sec:multiclass},  the size of the joint codebook can be decreased further by removing words with a low information gain. Figure~\ref{fig:ig_graph} shows the average classification performance as a function of the codebook size, which in turn depends on the threshold on information gain.  We can see from the figure that with information gain as a codebook selection criterion, the codebook size can be decreased more than 50\% with less than 5\% of classification accuracy loss.

\subsection{3D Detection}
\label{sec:results_depth}

We then evaluated the addition of the depth cue, described in Section \ref{sec:depth}. We assume that the changes in performance with and without depth are independent of the changes in performance with and without multi-class detection; a reasonable assumption given that the changes in data representation are themselves uncorrelated. Given this assumption, it is sufficient to evaluate the depth cue addition using training data containing a single class ($c_i=1$). 

A training dataset of side views of 40 different toy cars was collected with a Kinect$\textsuperscript{\textregistered}$ sensor (see Figure \ref{fig:3DTrainingExample}). Due to sensor limitations, an indoor setting was used. The variation in physical size was high in the dataset -- higher than among real cars. A version of the training set was therefore prepared, where all training images were scaled to the same width. (The test images were kept unscaled.)

For the first experiment below, a set of test images were collected, where cars from the training set were placed in different, cluttered environments, 
one car per image (see Figure \ref{fig:segEg} left). The test images are very challenging for three reasons: Firstly, the cars occupy a very small part of the images, secondly, the background is highly textured, and last, the cars vary in scale with more than a factor 2.

No car instance was present in both the training and test sets in the same run. Five training datasets were generated by randomly selecting training images of 30 toy cars for training, and cluttered test images of the remaining 10 unseen toy cars for testing. All experiments were then performed independently five times on the different datasets.

\begin{figure*}[t]
\centering
\subfigure[RGB image]{%
\includegraphics[width=0.3265\textwidth,height=43.5mm]{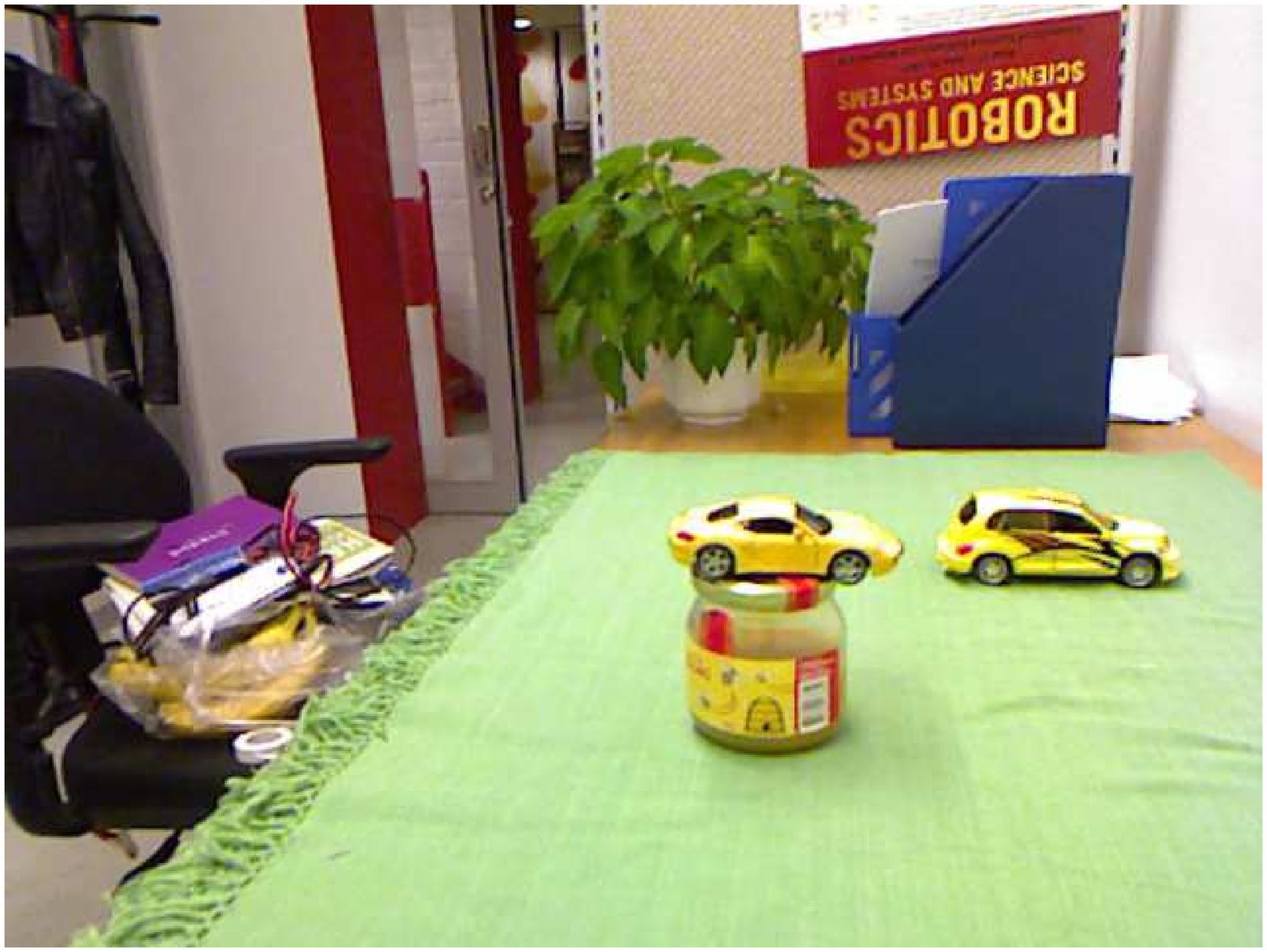}}
\subfigure[2D detection, baseline]{%
\includegraphics[width=0.3265\textwidth]{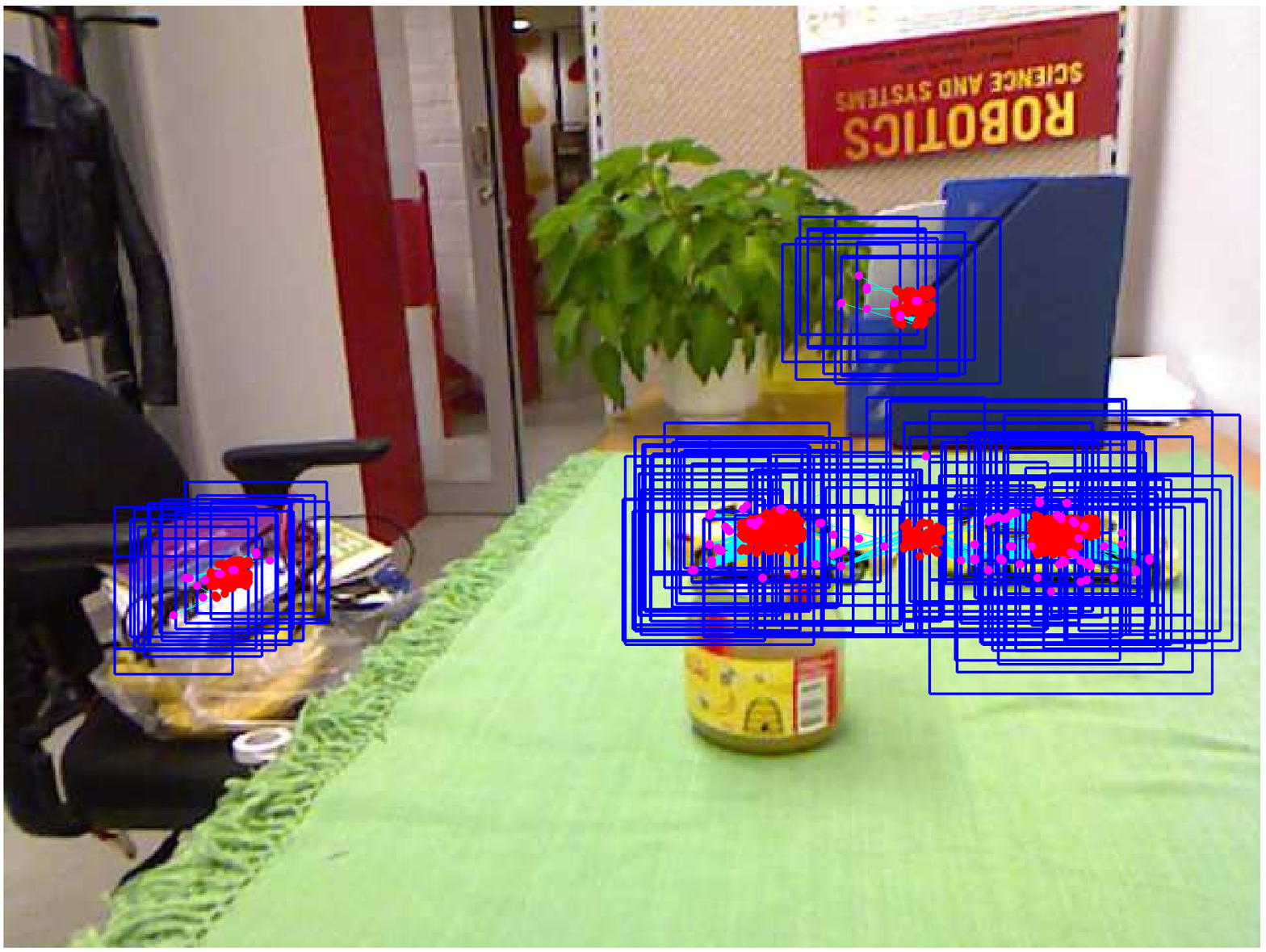}%
\label{fig:2D1}} 
\subfigure[2D segmentation, baseline]{%
\includegraphics[width=0.3265\textwidth]{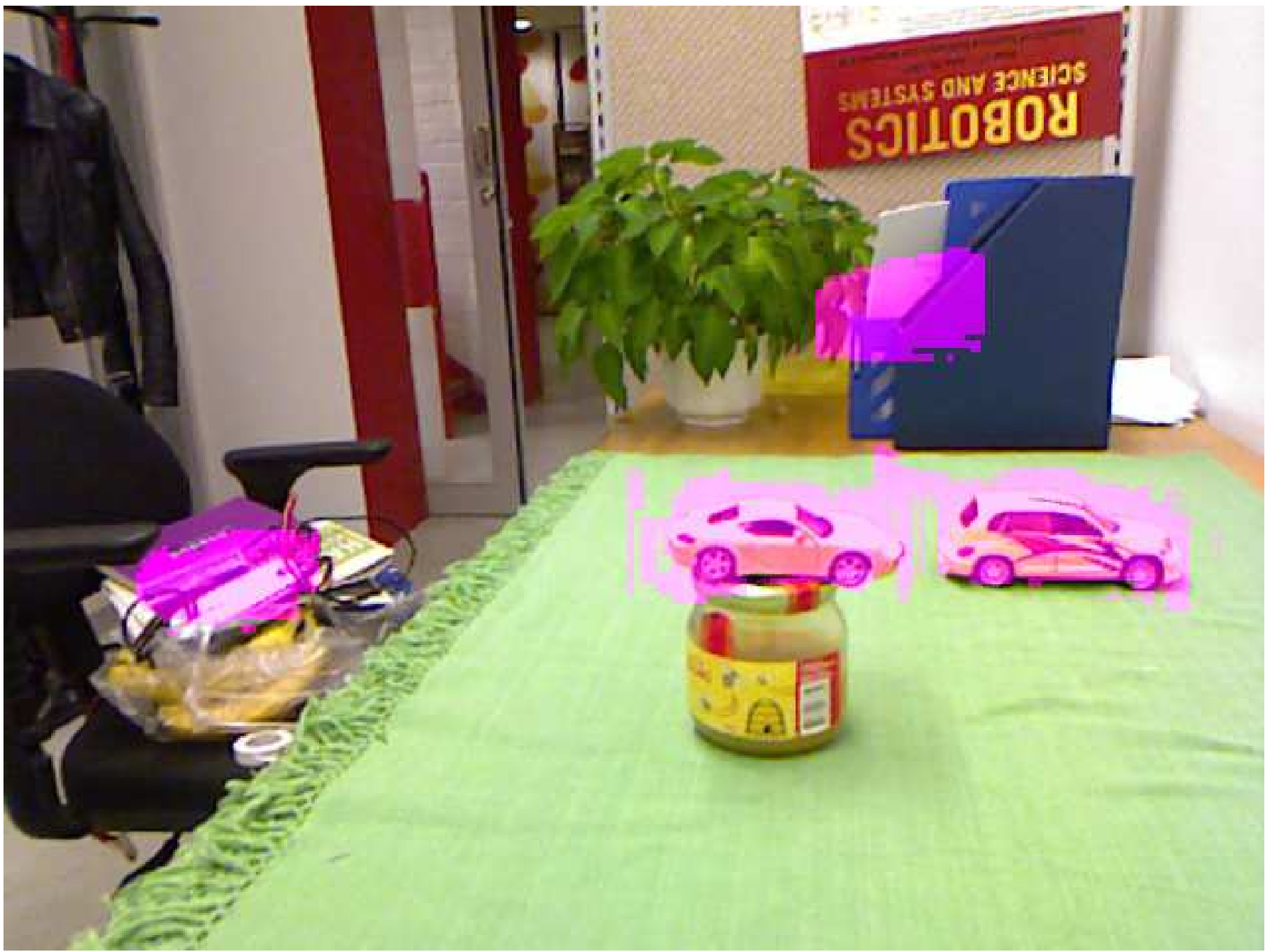}%
\label{fig:2D2}} 
\subfigure[Depth image]{%
\includegraphics[width=0.3265\textwidth]{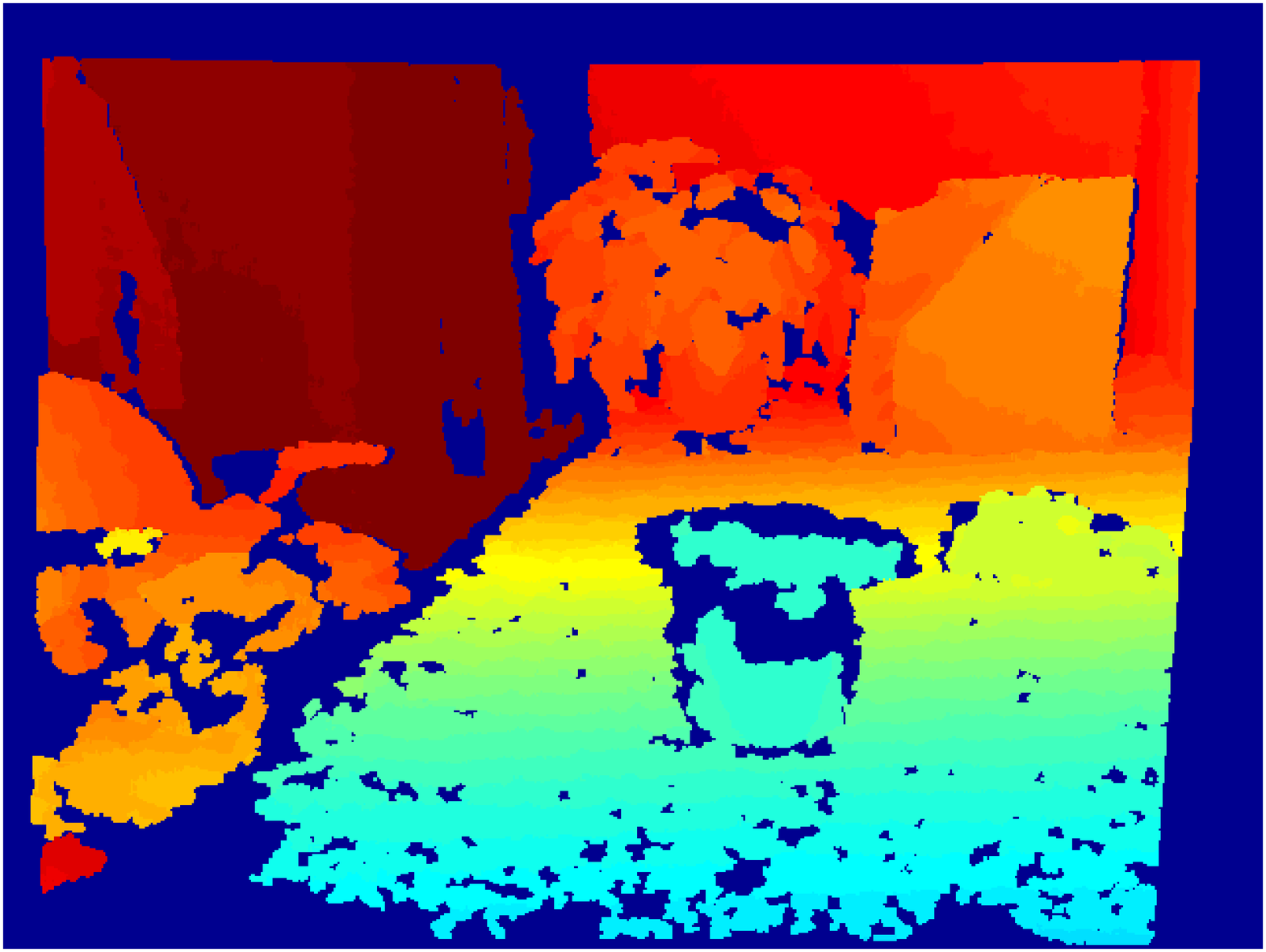}
\label{fig:sd3}} 
\subfigure[3D detection]{%
\includegraphics[width=0.3265\textwidth]{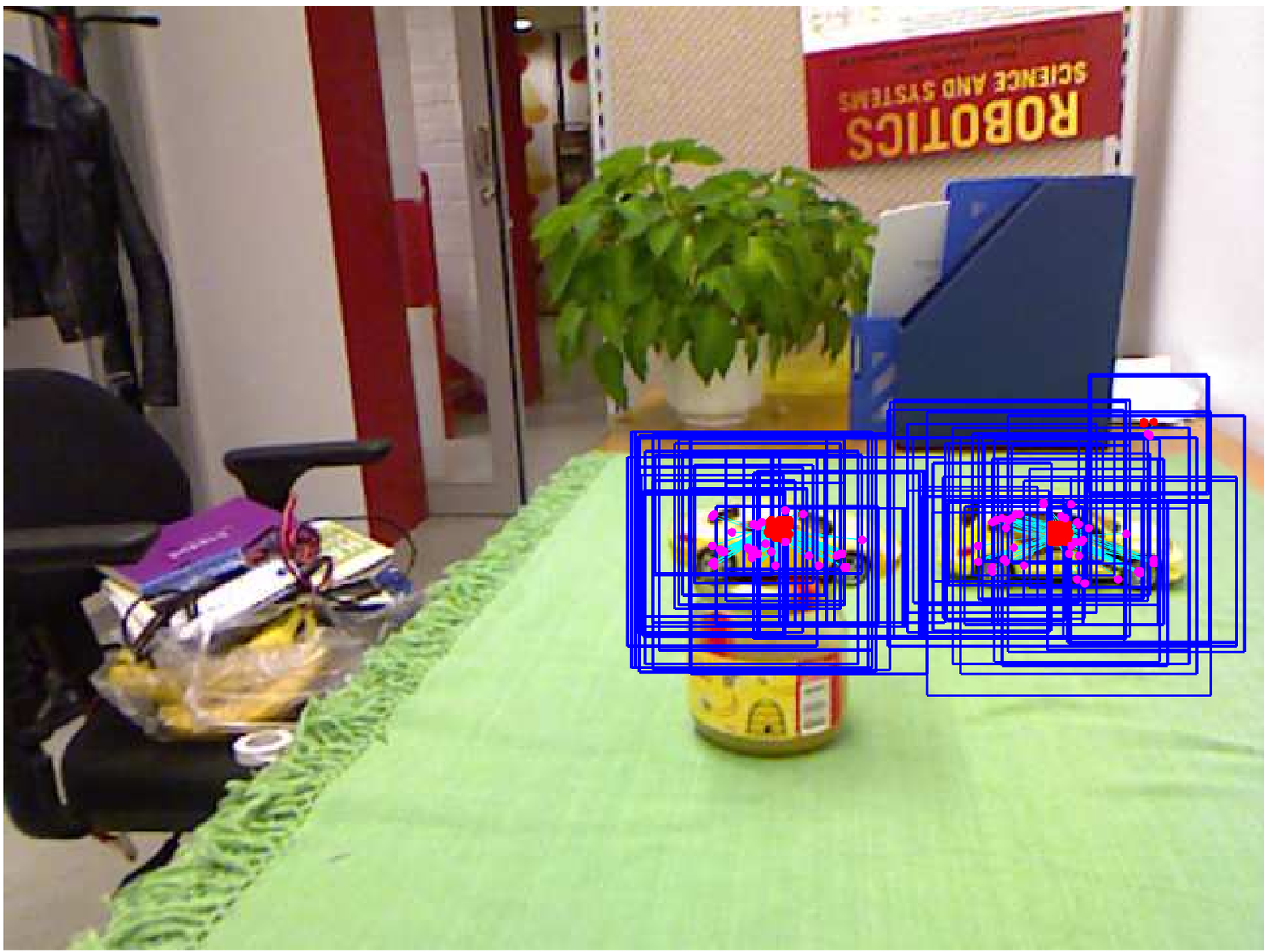}%
\label{fig:sd1}}
\subfigure[3D segmentation]{%
\includegraphics[width=0.3265\textwidth]{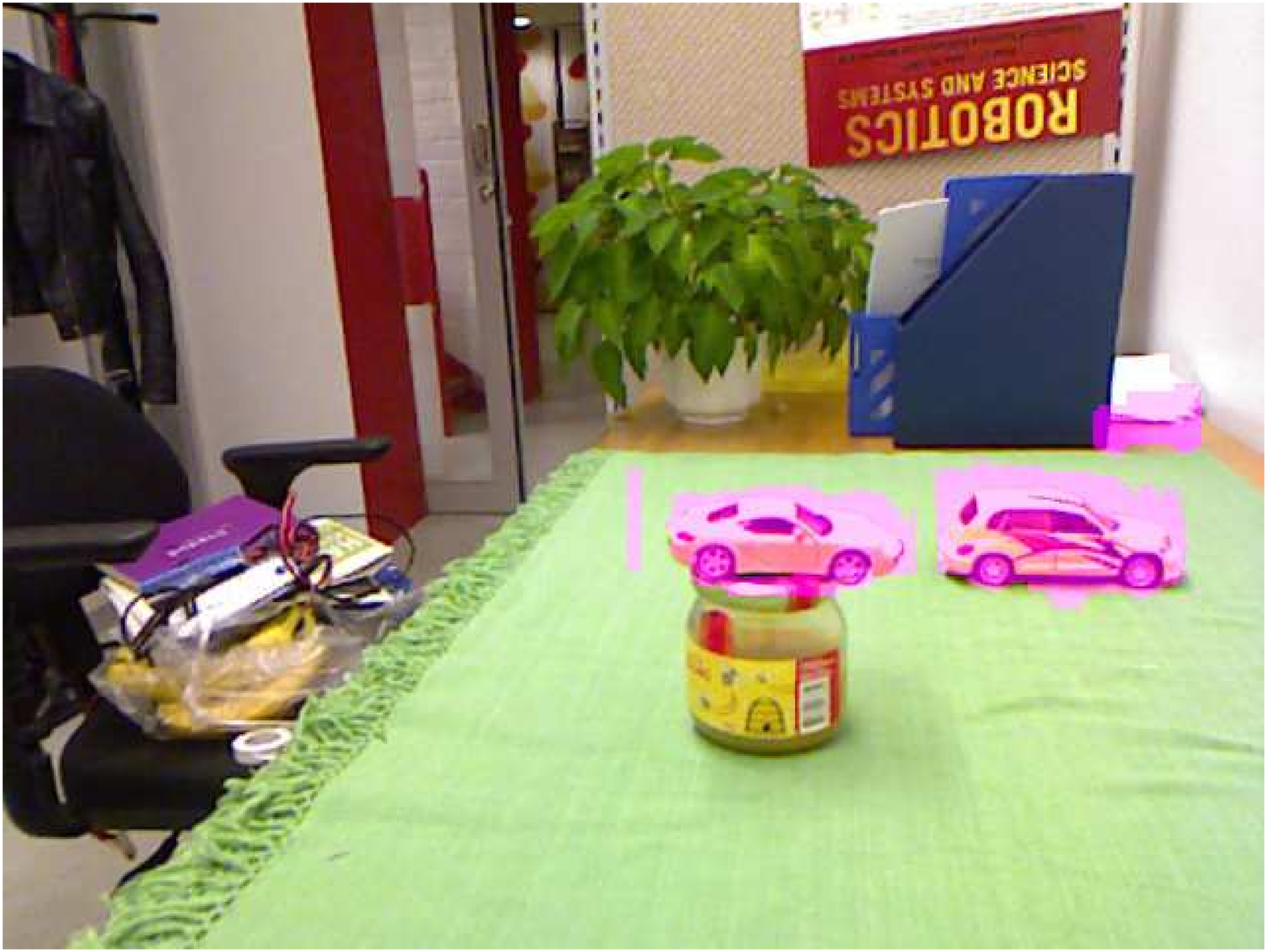}%
\label{fig:sd2}}
\caption{Example of multi-object detection. (b) and (c) show detection and segmentation using $V^\mathit{ISM}$ which does not employ depth information -- two correct detections and three false detections. (e) and (f) show the detection and segmentation using $V^\mathit{JI3SM\,3}$ -- two correct detections and one false detection. A smaller bandwidth was used than in Figure 9: 5 for $x$ and $y$, and 0.07 for $s\times d$}
\label{fig:2D_vs_sd}
\end{figure*}

A hypothesis brought forward in Section \ref{sec:depth} is that the depth cue increases the robustness to background noise. 
One way to measure this is to study the voting confidence, i.e., to what degree votes agree on the same object hypothesis.  
The results were:\\
With $V^\mathit{ISM}$ (Eq (\ref{eq:v-s})): \hspace{4mm} 58\%,\\
With $V^\mathit{JI3SM\,1}$ (Eq (\ref{eq:v-sandd})): 62\%,\\
With $V^\mathit{JI3SM\,2}$ (Eq (\ref{eq:v-d})): 64\%, \\
With $V^\mathit{JI3SM\,3}$ (Eq (\ref{eq:v-stimesd})): 66\%.\\ 
For voting, a bandwidth of 10 for $x$ and $y$, 0.01 for $s$, and 0.15 for $d$ and $s\times d$ was used.  We conclude that the voting confidence is indeed higher with depth cues, and that the $V^\mathit{JI3SM\,3}$ voting space gives the highest confidence.

The detection performance was then evaluated in terms of precision and recall. For this experiment, another test dataset was collected. Four car instances were randomly selected from the 40 cars, and the corresponding training images were removed from the training set (Figure \ref{fig:3DTrainingExample}). A series of 32 images with two or more test instances were then collected; for an example, see Figure \ref{fig:2D_vs_sd}(a,d). 

Detection in the images was carried out in the following way: Features were extracted, which voted for car hypotheses. Object hypotheses, i.e., local maxima in the hypothesis confidence space, were then detected. All hypotheses with a probability higher than $T_\mathrm{ratio}$=55\% of the probability of the strongest hypothesis were maintained as detections.
 
Figure \ref{fig:2D_vs_sd} shows an example detection result from this dataset; (e) showing the detection and (f) the segmentation of  $V^\mathit{JI3SM\,3}$, and as a baseline, (b) showing the detection and (c) the segmentation of $V^\mathit{ISM}$ which is not using depth.

The 2D detection in Figure \ref{fig:2D1} shows examples of the kind of erroneous detections that might occur when depth is not used. First of all, enough car-like spurious features were found on the plant, the blue folder and the chair, to give rise to two false detections. The features for both these detections are on different depths, which explains why they do not vote for the same hypothesis in the 3D detection in Figure \ref{fig:sd1}. Secondly, a more systematic error occurs: the two cars are aligned in the image so that the rear wheel of the right car supports the same hypothesis as the front wheel of the left car -- a car is "hallucinated" between the two real cars. However, since the cars are on significantly different depth (see Figure \ref{fig:sd3}), this false hypothesis does not occur if depth is taken into account (see Figure \ref{fig:sd1}).

\begin{figure}[t]
\centering
\includegraphics[width=\linewidth,height=64mm]{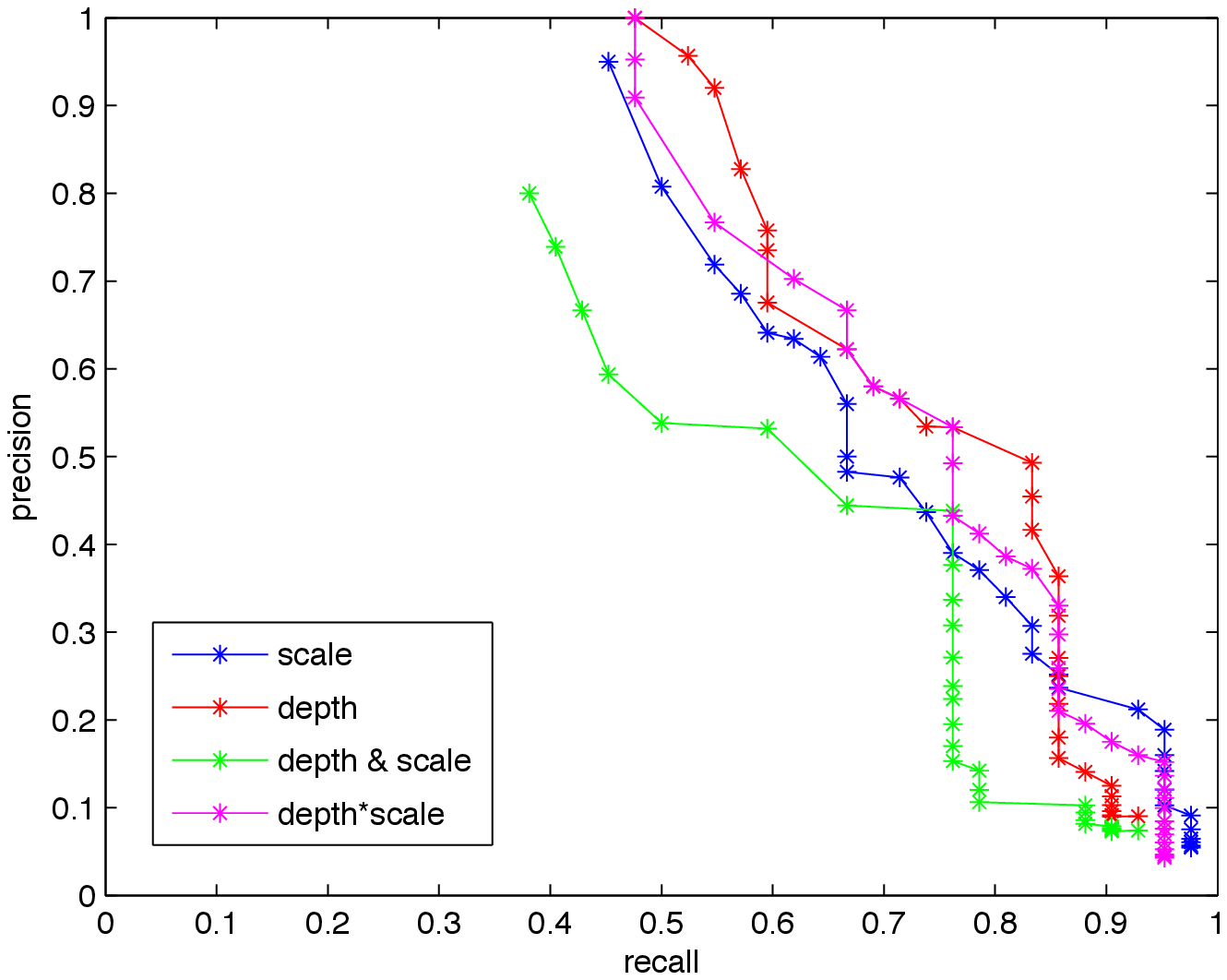}
\caption{Precision-recall curves for detection using the $V^\mathit{ISM}$ (blue), $V^\mathit{JI3SM\,1}$ (green), $V^\mathit{JI3SM\,2}$ (red), and $V^\mathit{JI3SM\,3}$ (magenta) voting spaces.}
\label{fig:compareD}
\end{figure}

Figure \ref{fig:compareD} gives the precision-recall curves of the first stage detection result using the four alternative voting spaces. The curves were generated by varying the threshold ($T_\mathrm{ratio}$) of the ratio between the probability of detections and that of the strongest detection in the image. This threshold controls the recall factor as:  $(T_\mathrm{ratio} \rightarrow 0) \Leftrightarrow (\mathrm{recall} \rightarrow 1)$.

The curves confirm that using depth in the detection gives a more stable performance than detection with 2D cues only. However, increasing the dimensionality of the voting space, as in $V^\mathit{JI3SM\,1}$ (green curve) gives a significantly worse detection performance. The reason for this is most certainly that the JI3SM model with the larger space requires more training data -- the current training set of 36 cars is simply too small for the model to converge. Future work includes evaluation with a larger training set (Section \ref{sec:conclusions}).

\section{Conclusions}
\label{sec:conclusions}

We presented a 3D, multi-class object detection and segmentation method, intended for a humanoid robot perception application. To that end, we extended the 2D, single-class detection method of Leibe et al.~\cite{Erdos01} to handle multiple classes using a joint minimal codebook, and to incorporate depth measurements to enhance the robustness of the voting procedure of the detection step. Both these extensions are essential for the method to be of use on a humnoid robot functioning in human environments.  Moreover, we still kept all the advantages of the original method, e.g., rough pose estimation by learning classes corresponding to both viewing direction and object class.
 
The experiments showed that with the new multi-class model, the same detection accuracy could be obtained as with a set of single-class models, but with a gain in codebook size:  the codebook size could be lowered to half with less than 5\% detection accuracy loss.  Moreover, it was shown that the introduction of a depth cue in the method improved detection performance, in that votes from spurious background features and other objects were filtered out more efficiently in the object detection stage.  

As discussed in Section \ref{sec:depth}, depth is not currently used for segmentation explicitly. However, there is rich information in the range image which could be exploited for that purpose. Using shape features \cite{Erdos62,Erdos63} together with visual features (e.g., SIFTs) would most certainly increase both the detection and 3D segmentation performance. 

We also intend to integrate the current method on a real humanoid robot, and further evaluate the performance of 3D detection and segmentation of different classes of indoor objects. This involves collecting a database of RGB-D images of a large number of object classes, including several instances of each class. 

Another avenue of research to explore further is that of more elaborate category models. When the number of classes grow, the classification is improved by introducing structure, e.g., topic models such as pLSA \cite{Erdos65}, LDA \cite{Erdos64} or DiscLDA \cite{Erdos53}. Such hierarchical and structural models are also more suited for reasoning about objects in grasping and manipulation applications, where topics or features can be correlated to robot grasping strategies \cite{Erdos66}.

\bibliographystyle{IEEEtran}
\bibliography{ref}
\end{document}